\documentclass[a4paper,fleqn,11pt]{article}

\usepackage[utf8]{inputenc}
\usepackage{amsfonts}
\usepackage{apacite}

\usepackage{pslatex}
\usepackage{graphicx,color}
\usepackage{amsmath}

\newtheorem{mydef}{Definition}

\newtheorem{myhyp}{Proposition}

\title{The Tensor Brain: \\ Semantic Decoding for Perception and Memory}

\author{{\large \bf Volker Tresp$^{1,2}$\footnote{Corresponding author, Volker.Tresp@lmu.de}, Sahand Sharifzadeh$^{1}$, Dario Konopatzki$^{1,2}$, Yunpu Ma$^{1,2}$ } \\
$^{1}$Ludwig Maximilian University of Munich; $^{2}$Siemens, Corporate Technology, Munich}

\date{}

\begin{document}

\maketitle

\begin{abstract}

%
%

We analyze perception and memory, using mathematical models for knowledge graphs and tensors, to gain insights into the corresponding functionalities of the human mind. Our discussion is based on the concept of propositional sentences consisting of \textit{subject-predicate-object} (SPO) triples for expressing elementary facts. SPO sentences are the basis for most natural languages but might also be important for explicit perception and declarative memories, as well as intra-brain communication and the ability to argue and reason. Due to its compositional nature, a set of sentences can describe a scene in great detail, avoiding the explosion in complexity with flat representations. A set of SPO sentences can be described by a knowledge graph, which can be transformed into an adjacency tensor. We introduce tensor models, where concepts have dual representations as indices and associated embeddings, two constructs we believe are essential for the understanding of implicit and explicit perception and memory in the brain. We argue that a biological realization of perception and memory imposes constraints on information processing. In particular, we propose that explicit perception and declarative memories require a complex semantic decoder, which, in a basic  realization, has  four layers:
First, a \textit{sensory memory layer}, as a buffer for sensory input, second, a memoryless \textit{representation layer} for the broadcasting of information ---the ``blackboard'', or the ``canvas'' of the brain---, third,
an \textit{index layer} representing concepts, and fourth, a \textit{working memory layer} as a processing center and data buffer. We discuss the operations of the four layers and relate them to the global workspace theory. Whereas simple semantic decoding might be performed already by higher animals, the generation of triple statements, requiring working memory as part of a complex semantic decoder, is a layered sequential process likely performed only by humans. In the resulting chatterbox decoding, semantic consistency is encouraged on the representation level. Both semantic and episodic memory contribute context and thus complement sensory input with non-perceptual information: agents have memory systems for a purpose, i.e., to make better decisions! In a Bayesian brain interpretation, semantic memory defines the prior distribution for observable triple statements. We propose that ---in evolution and during development--- semantic memory, episodic memory, and natural language evolved as emergent properties in agents' process to gain a deeper understanding of sensory information. Our mathematical model provides some fresh perspectives on much-debated issues concerning the relationship between perception, semantic memory, and episodic memory. We present a concrete model implementation and validate some aspects of our proposed model on benchmark data where we demonstrate state-of-the-art performance.

\end{abstract}

\section{Introduction}

With an increase in higher animals' abilities to move and act came a growing demand for high-performing perceptual systems \cite{hommel2001theory}. We argue in this paper that, in the course of this development, it became an important faculty that an agent could analyze relationships between entities, e.g., in visual scenes, and that for this faculty, the agent needed a complex semantic decoder, i.e., a mapping from a sensory vector space to explicit sentences describing, e.g., a visual scene.

By achieving complex semantic decoding, agents developed the ability to communicate, e.g., they could inform
other agents that there is a leopard waiting outside the hide-out. Further development included the ability to argue, and human agents even developed the ability for logical reasoning, which eventually enabled humans to explore mathematics and the sciences.

Whereas simple semantic decoding might be a capability of higher animals, we propose that complex semantic decoding requires working memory and might be a faculty unique to humans. The ability to perform complex semantic decoding
might also be the reason why humans are capable of grammatical language, beyond the simple utterances of other primates.

An agent needs more than just perception:
it needs to remember where it had been before,
why it is where it is,
and what the general context is, beyond the here and now.
For example,
an agent needs to remember that even though perception does not give a clue,
there is still a leopard waiting outside the hide-out.
Thus there is a need for remembering the immediate past (the agent remembers that it is in the hide-out because the leopard was chasing it),
and this ability developed into episodic memory.
Once developed, several side benefits evolved.
Episodic memory can provide training data for the brain, it reminds the agent about past situations similar to the current one, and it contributes to imagination and planning \cite{kumaran2016learning}.

Another faculty agents developed is semantic memory, which consists of two parts. First, there
is knowledge about which concepts in the world exist, like classes of things, entities, locations, attributes, and predicates.
Second, there is knowledge of facts concerning these concepts.
 Semantic memory can be important for survival: humans simply know that lions are dangerous, even if a lion looks cozy and sleepy and even if an individual did not yet have an unpleasant encounter with a lion.
 One can say that episodic memory is more egocentric (what did the agent experience), whereas semantic memory is more allocentric (what is true, independent of the personal experience of the agent), although as we argue, semantic memory, to a great extent, also evolved out of an agent's perception.
 We discuss the close relationship between semantic memory (``things we know'') and episodic memory (``things we remember'').
 We suggest  that semantic memory might have emerged as a by-product of perception and its semantic decoding.
 We propose that semantic memory does not only provide world knowledge, it also defines the prior distribution for triple statements.

In this paper, we focus on sentences in the form of
\textit{subject-predicate-object} (SPO) triples.
The first reason is that, in most languages, basic facts are expressed in SPO format and thus triple statements are arguable of fundamental relevance.
A second reason is that a whole knowledge base of SPO sentences can be described as a knowledge graph, where the entities in the sentences are nodes and predicates become labeled directed links pointing from subject node to object node.
Knowledge graphs are universal in the sense that statements involving predicates with an arity larger than two can be reduced to triple formats.


Knowledge graphs currently have a large impact in applications and industry.
The most prominent example is the Google Knowledge Graphs with currently on the order of 100B triples \cite{singhalintroducing2012}.
Further popular large-scale knowledge graphs are DBpedia \cite{auerdbpedia:2007}, YAGO \cite{suchanekyago:2007}, Freebase \cite{bollackerfreebase:2008}, and NELL \cite{carlsontoward2010}.

Knowledge graphs can be described by their adjacency tensors. Due to their large scale,
it is infeasible to directly work with tensors, but one can work with tensor models, instead \cite{nickel2015}.
We define tensor models to be mappings from indices to index embeddings, which are then mapped to real numbers.
We believe that indices and their embeddings are essential concepts for understanding the working of the brain.

We propose that an embedding-based implementation of
 tensor models for perception and memory, under some constraints imposed by biology, requires four important layers.
 First, the \textit{sensory memory layer}, which is a buffer that maintains a representation of scene information.
 Second, the \textit{representation layer}, which is the brain's main communication platform (``theater of the brain''). Third, the \textit{index layer}, which indexes the individual's acquired concepts and time instances. Finally, complex semantic decoding requires a memory function, which is supplied by the \textit{working memory layer}. The model can realize perception, including the generation of triple statements for scene descriptions,
 as well as episodic and semantic memory.

 From a biological point of view, the representation layer might be part of the posterior hot zone, which, as it as been argued, is the minimal neural substrate essential for conscious perception \cite{koch2016neural}. In conjunction with working memory,
one might relate it to the prefrontal parietal network (PPN), which has been argued to be a
basis for consciousness \cite{bor2012consciousness}, and the global workspace \cite{baars1997theater,dehaene2014consciousness}.

Complex semantic decoding of perceptual input and memory thus is a layered sequential process, in which the embeddings of decoded concepts are communicated to the brain, as a whole.
A set of decoded triple statements, including triples from semantic memory and episodic memory, might lead to an integrated representation that
 permits refined decision making by the agent: The agent does not only understand better --- due to memory recall, it also acts better!
 We analyze how memories can be queried by a stochastic sampling process, which leads to a generative model, we actually propose to be implemented in biology.
 In operation, we produce sampled triples. The importance of sampling has been discussed by \cite{dehaene2014consciousness} in the context of conscious perception.
 For example, he states that ``... consciousness is a slow sampler''.

The paper is organized as follows.
In the next section, we cover related work.
In Section~\ref{sec:KGs} we introduce knowledge graphs, their adjacency tensors, and tensor models.
We distinguish between the common static knowledge graph,
the temporal knowledge graph,
and the probabilistic knowledge graph. 
In Section~\ref{sec:serial} we discuss a stochastic serialization of knowledge graphs by a sampling process.
Section~\ref{sec:disc1} presents some modeling details and discusses supervised and self-supervised learning.
In Section~\ref{sec:bioImp} we discuss an architecture that obeys some constraints imposed by information processing in the brain.
A reader only interested in the actual model might directly jump to this section.
 The architecture assumes the form of a special recurrent neural network and actual implementations are described in the appendix.
In Section~\ref{sec:discussion} we relate our model to current discussions in cognition and neuroscience. We also extend our model to include an episodic memory.
In Section~\ref{sec:exp} we describe experimental results using benchmark data sets.
Section~\ref{sec:concl} contains our conclusions.

\section{Related Work}

\subsection{Tensor Models for Knowledge Graphs}

RESCAL was the first tensor-based embedding model for triple prediction in relational data sets and knowledge graphs \cite{nickel2011three,nickelfactorizing2012}. Before that, tensor factorization was applied to retrieval and ranking \cite{franztriplerank:2009}. Embedding learning for knowledge graphs evolved into a sprawling research area \cite{bordestranslating2013,socherreasoning2013,yang2014embedding,nickel2015holographic,trouillon2016complex,dettmers2018convolutional}. \cite{nickel2015} provides an overview.

\subsection{Cognitive Tensor Models and Related Models}

The line of work described in this paper started with \cite{tresp2015learning}. That paper introduced
tensor models with index embeddings for perception, as well as
 semantic and episodic memory. The paper does not contain a biologically plausible implementation of the tensor models and also does not contain experiments.

 \cite{tresp2017tensor, tresp2017ccn,tresp2017embedding, ma2018embedding} analyzed the connection between temporal, episodic and semantic tensor models.
 In those papers, the temporal knowledge graph was modeled by a Bernoulli likelihood function, from which semantic memory was derived by an integration  step, performed in latent space.
 In this paper, we define models for conditional categorical probabilities, which permits the efficient sampling of triples and also the integration of visual input.

 In 2016, the Stanford Visual Relationship dataset was published which contained images annotated with triple statements \cite{lu2016visual}. \cite{baier2017improving} showed how a prior distribution derived from triple occurrences could significantly improve on pure vision-based approaches and on approaches that used prior distributions derived from language models.
 \cite{sharifzadeh2019improving} showed further improvements by including 3-D image information.
 \cite{trespmodel2019} is a short paper that describes an earlier version of the model described in this paper.

Tensor decomposition has been used previously as memory models but the main focus was on simple associations \cite{hintzman1984minerva,kanerva1988sparse,humphreys1989different,osth2015sources} and compositional
structures \cite{smolensky_tensor_1990,pollack1990recursive,platecommon1997,halfordprocessing1998,ma2018holistic}. In the tensor product approach \cite{smolensky_tensor_1990}, encoding or binding is realized by a tensor product (generalized outer product),
and composition by tensor addition. In the STAR model \cite{halfordprocessing1998}, predicates are represented as tensor products of the components of the predicates.
None of these
approaches define an optimization step for tensor factorization but have some form of factor design,  instead, e.g., by using random vectors as embeddings.

Recent developments \cite{schlag2018learning} have shown that tensor product representations with learnable embeddings, instead of random embeddings, can be profitably combined with recurrent neural networks to express the combinatorial representation of sequential data. Further developments of the tensor product approach demonstrated that tensor product decomposition networks with trainable filler-role pairs can improve on recurrent neural network encoders \cite{mccoy2018rnns}.

 \subsection{Scene Graphs}

Triple statements generated from an image form a scene graph \cite{johnson2015image}. Work on scene graphs attempts to find a unique, and globally optimal, interpretation of an image. The modern work on scene graphs started with \cite{lu2016visual} and \cite{krishna2017visual}. The two papers made their annotated data available, which spawned an explosion in research. The first papers addressed visual relation detection \cite{baier2017improving,zhang2017visual,baier2018improving}. State-of-the-art scene graph models are
described in \cite{yang2018graph,zellers2018neural,hudson2019learning}.
The background information in \cite{lu2016visual} was extracted from a text corpus. Recent work in this direction is \cite{luo2019context}.

In the work presented in this paper, the focus is on chatterbox decoding, which produces a
 set of triples. Global consistency
is reflected in shared representations.
Arguments for the chatterbox approach are that, first, consistency on the level of the extracted triple statements could be addressed by a later processing stage,
 second, the agent's real-life video might change too fast to permit the analysis of a static scene graph, and third,
 biologically plausible implementations of scene graph models would add a lot of model complexity. The focus of this paper is on the simplest possible models satisfying our biological and technical constraints. By chaining the decoding, dependencies between triples can further be increased.
 Scene graphs are discussed again in Section~\ref{sec:discussion}.

\subsection{Related Modern Technical Models for Memory}

\cite{hochreiter1997long} convincingly demonstrated the importance of memory systems in recurrent neural networks. Important later extensions were the
neural Turing machine (NTMs) \cite{graves2014neural}
and the memory networks \cite{weston2014memory}. In those approaches, episodic memory acts as a memory buffer for training recurrent neural networks and for reinforcement learning.

\subsection{Dual Process Theory}

In psychology, dual process theory concerns the interplay in the mental processing of an
 implicit, automatic, unconscious process (shared with animals) and an explicit, controlled, conscious process (uniquely human).
 See \cite{evans2003two} for a review. One instance is Kahneman's system-1 / system-2 dichotomy \cite{kahneman2011thinking}.

CLARION is a dual-process model of both implicit and explicit learning \cite{sun1996learning}.
It is based on one-shot explicit rule learning (i.e., explicit learning) and gradual implicit tuning (i.e. implicit learning).

In our model, the implicit side would be on the level of embeddings and representations, whereas the
explicit side is on the level of the concept indices and the extracted triple statements.

A recent approach is \cite{bengio2017consciousness}. Although it addresses similar issues as our work,
the approach is quite different with a focus on explicit sparse factor graphs. See the discussion in Section~\ref{sec:discussion}.

\subsection{The Bayesian Brain}

Our approach is in the
tradition of Bayesian approaches to brain modeling
 \cite{dayan1995helmholtz,rao1999predictive,knill2004bayesian,kording2004bayesian,tenenbaum2006theory,griffiths2008bayesian,friston2010free}.

In \cite{baier2017improving} an explicit semantic prior distribution was used, describing \textit{a priori} probabilities for triple statements. For inference, Bayes' formula is used.
The great improvement in performance after integrating the prior information is an indication that triple representations might be a powerful abstraction level for formulating prior information.

In the approach presented here, we assume that semantic memory defines a prior distribution which is conjugate to the implicit likelihood function, in the sense that the posterior sampling distribution and the prior sampling distribution can be described by the same basis. We do not explicitly apply Bayes' formula. Instead, we have a shared representation for prior and posterior distributions.

\section{Knowledge Graphs and their Adjacency Tensors}
\label{sec:KGs}

\subsection{The Static Knowledge Graph $\textit{KG}_{\textit{static}}$}

\begin{figure}[t]
\vspace{-1cm}
\begin{center}
\includegraphics[width=0.8\linewidth]{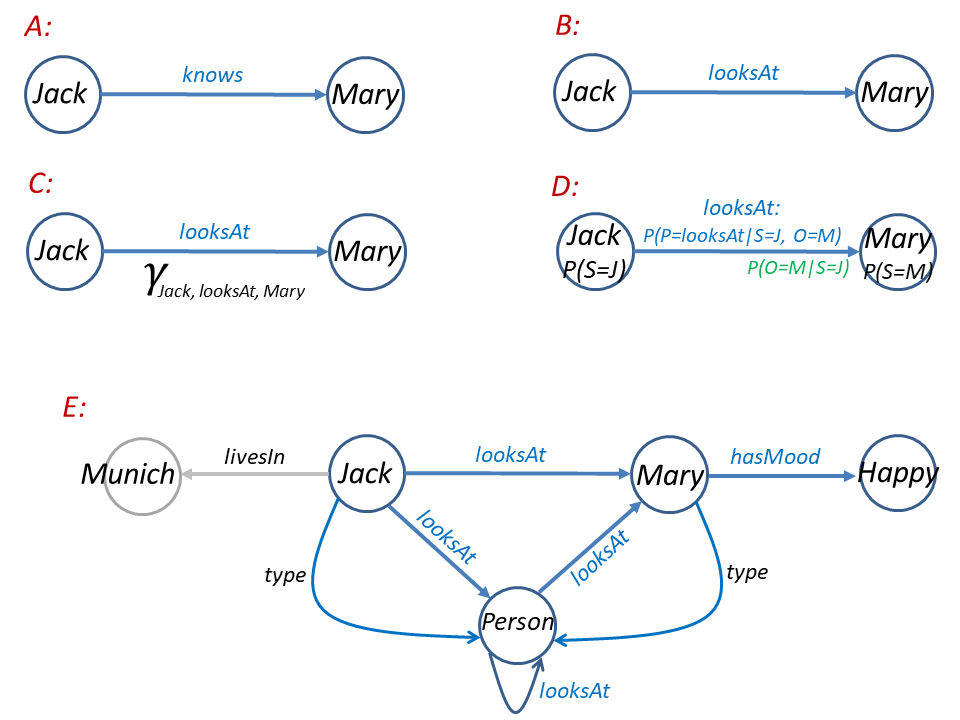}
\end{center}
\vspace{-0.5cm}
\caption{
\textbf{A:} A triple in the static KG, which states that {Jack knows Mary}.
\textbf{B:} A triple of the temporal KG. {Jack looks at Mary} at a time instance $t$.
\textbf{C:} A triple of the probabilistic KG. {Jack looks at Mary} at any time with
probability $\gamma_{\textit{Jack, looksAt, Mary}}$.
 \textbf{D:} Triple generating process for the probabilistic KG: Jack is the subject with probability $\mathbb{P}(S=J)$. Mary is the subject with probability $\mathbb{P}(S=M)$.
Mary is the object when Jack is the subject with probability $\mathbb{P}(O=M|S=J)$.
If Jack is the subject and Mary is the object, then the probability that Jack looks at Mary is $\mathbb{P}(P=looksAt | S=J, O=M)$.
 \textbf{E:} Same but with an additional nonvisual predicate ``livesIn'' and nodes for
entities (\textit{Jack, Mary}), a class (\textit{Person}), a location (\textit{Munich}), and an attribute (\textit{Happy}).
}
\label{fig-KG}
\end{figure}

We consider a world consisting of a set of $N_E$ entities $\mathcal{E} = \{e_1,\dots, e_{N_E}\}$ and
a set of $N_P$ relation types or predicates ${\mathcal{P} = \{p_1, \dots, p_{N_P}\}}$.\footnote{Entities can, e.g., be persons, objects, locations. Later in the discussion, entities will stand for general concepts.} We are interested in triple statements of the form
$(s, p, o)$ where $s \in \mathcal{E}$ is the identifier for the subject entity, $p\in \mathcal{P}$ is the identifier for the predicate,
 and
$o\in \mathcal{E}$ is the identifier for the object entity.
Examples for triple statements are:
\textit{(Munich, partOf, Bavaria)},
\textit{(AkiraKurosawa, directorOf, SevenSamurai)},
 \textit{(Jack, knows, Mary)},
 \textit{(Jack, type, Person)},
 \textit{(Jack, height, Tall)},
 and \textit{(Jack, age, Young)}.
 In a static knowledge graph, $\textit{KG}_{\textit{static}}$, the entities are the nodes and a true triple statement
 is indicated by a directed link from $s$ to $o$, labeled by $p\in \mathcal{P}$. See Figure~\ref{fig-KG}A.
We write, $\textit{KG}_{\textit{static}} \subseteq {\mathcal{E}} \times {\mathcal{P} } \times {\mathcal{E}}$.
 We can describe
$\textit{KG}_{\textit{static}}$ as a third order tensor:
\begin{mydef}
The triples in the {static knowledge graph} $\textit{KG}_{\textit{static}}$ can be grouped as a
third-order adjacency tensor
${\underline{\mathbf{Y}}_{\textit{static}}} \in \{0,1\}^{N_E \times N_P \times N_E}$,
whose entries are set such that
\[
y_{s, p, o} = \begin{cases}
 1,& \text{if the triple } (s, p, o) \text{ exists} \\
 0, & \text{otherwise.}
\end{cases}
\]
It is a function,
$\mathcal{E} \times \mathcal{P} \times \mathcal{E} \rightarrow \{ 0, 1 \}$.
\end{mydef}
In applications, the majority of currently employed KGs are static KGs.  Since a static KG is about facts that are either true of false, we consider it a Boolean KG.

But what is indeed true?
We assume that there is a ground truth oracle which returns true, or $1$, for all true triple statements and otherwise $0$.
True triple statements can have a scientific character, like ``Earth revolves around the sun'' or human conventions like ``Munich is part of Bavaria''. For practical purposes, we might consider a true sentence as a statement where the majority of domain experts would agree that it is true.
 Triple statements where the oracle returns a $0$ are considered false,
as ``Munich is part of Belgium'' (closed-world assumption) or as triples with unknown truth values (open-world assumption).
We exclude statements that are ill-posed or cannot be formulated in triple format.
This restriction on simple true facts made knowledge graphs a success in many applications.


The static KG can be related to a brain's semantic memory, since it is about facts, an agent assumes to be true.

\subsection{The Temporal Knowledge Graph $\textit{KG}_{\textit{temp}}$}


Most knowledge graphs, which have been developed in the past, are static and reflect triple statements which are stable in time. In reality, of course, the state of the world is changing: a healthy
person becomes diagnosed with a disease and a new president is inaugurated. This can be described by temporal knowledge
graphs. See Figure~\ref{fig-KG}B.

Consider time instances $\mathcal{T} = \{t_1,\dots, t_{N_T}\}$. The KG at time instance $t \in \mathcal{T}$, $\textit{KG}_t$, contains the set of all triples that are true at time instance $t$ and we write $(s, p, o)_t \in \textit{KG}_t$,
 with $t = 1, \ldots, N_T$.
Thus, $\textit{KG}_t$ describes the state of the world at time instance $t$.
 Whereas the static KG would formulate that \textit{(s, p, o)} is a true sentence independent of the time instance, the temporal KG would formulate that \textit{(s, p, o)} is a true sentence
 at time instance $t$.

The temporal knowledge graph $\textit{KG}_{\textit{temp}}$ is simply the set of all $\textit{KG}_t$.
$\textit{KG}_{\textit{temp}}$ can display different temporal patterns. ``Surprising'' patterns are singular events (``signing of a peace treaty''). An event might indicate a state transition, e.g., before
the signing of a peace treaty (the event), a country was at war, and afterward, it was at peace.
Memorable events play a particular role in episodic memory.
Thus, in addition to static triples, the temporal $\textit{KG}_{\textit{temp}}$ at time instance $t$ contains triples, e.g., for the singular events
\textit{(Jack, diagnosed, Diabetes)},
\textit{(Jack, getsInjection, MeaslesVaccine)},
 \textit{(Jack, looksAt, Mary)}.

\begin{mydef}
 The triples in the {temporal knowledge graph} $\textit{KG}_{\textit{temp}}$ can
be grouped as a
fourth-order adjacency tensor
${\underline{\mathbf{Y}}_{\textit{temp}}} \in \{0,1\}^{N_E \times N_P \times N_E \times N_T}$,
whose entries are set such that
\[
y_{s, p, o, t} = \begin{cases}
 1,& \text{if the triple \textit{(s, p, o)} is true at time instance $t$} \\
 0, & \text{otherwise.}
\end{cases}
\]
It is a function,
$\mathcal{E} \times \mathcal{P} \times \mathcal{E} \times \mathcal{T} \rightarrow \{ 0, 1 \}$.
\end{mydef}
A temporal KG is also a Boolean KG. An agent's $\textit{KG}_{\textit{temp}}$ could be related to an ideal episodic memory, which can recover the state of the world, respectively significant events, at any time in the past (see Section~\ref{sec:discussion}).

\subsection{
The Probabilistic Knowledge Graph $\textit{KG}_{\textit{prob}}$}

In a simple generative model for $\textit{KG}_{\textit{temp}}$, we assume that the truth values of triple statements are generated by independent Bernoulli distributions.
Thus, with $0\le \gamma_{s,p, o} \le 1$,
\begin{equation}\label{eq:expsem1}
\mathbb{P}(y_{s, p, o, t}=1) = \gamma_{s,p, o} .
\end{equation}
 If
$\gamma_{s,p, o} = 1$, the triple is known to be always true (and would be part of the static KG), if $\gamma_{s,p, o} = 0$, the triple is known to be never true, and if
$0<\gamma_{s ,p, o}<1$, then $\gamma_{s,p, o}$
indicates the proportion of times that the triple is known to be true. An example for the first case would be \textit{(Munich, locatedIn, Bavaria)}, the second case \textit{(Munich, locatedIn, Netherlands)}, and the last case,
\textit{(Munich, hasTemperature, Hot)}.

 We can introduce the {probabilistic knowledge graph}, $\textit{KG}_\textit{prob}$, where we attach the weight $\gamma_{s,p, o} $
 to the labeled links (Figure~\ref{fig-KG}C). We can define the corresponding adjacency tensor:
\begin{mydef}
The triples in the \emph{probabilistic knowledge graph} $\textit{KG}_{\textit{prob}}$ can be grouped as a
 third-order adjacency tensor
${\Gamma}_{\textit{prob}}$
with entries $\gamma_{s, p, o}$. We refer to this as the
 \textit{adjacency tensor for $\textit{KG}_{\textit{prob}}$}.
 It is a function,
$\mathcal{E} \times \mathcal{P} \times \mathcal{E} \rightarrow [0, 1]$.
\end{mydef}

%

Equation~\ref{eq:expsem1} describes a generative model for the temporal knowledge graph $\textit{KG}_\textit{temp}$.
Thus in our modeling assumption, $\gamma_{s, p, o}$ is the probability that $y_{s, p, o, t} = 1$
prior to (or without) any observation at time instance $t$.
{By defining a prior distribution, we can also interpret $\gamma_{s, p, o}$ as the agent's personal belief that the
triple statement \textit{(s, p, o)} is true in the next instance in time.

Since $\gamma_{s,p, o} \approx \mathbb{E}_t (y_{s, p, o, t})$, it is clear that we can obtain an estimate of the
probabilistic KG adjacency tensor as an average of the temporal KG adjacency tensor.

As discussed in Section~\ref{sec:discussion}, the probabilistic KG can also be related to a brain's semantic memory.

\subsection{Tensor Models with Index Embeddings}

To achieve scalable solutions, and to improve the accuracy of estimates, we employ parameterized models.
Standard tensor decompositions are described by multilinear maps.\footnote{An example for a multilinear tensor factorization model is RESCAL \cite{nickel2011three}, with
$
\textit{f}^{\textit{prob}}_{\mathbf{w}}(\mathbf{a}_s, \mathbf{a}_p, \mathbf{a}_o)
= \mathbf{a}_o^T \textit{matrix}(\mathbf{a}_p) \mathbf{a}_s
$
where $\textit{matrix}(\cdot)$ transforms the vector into a square matrix.
Multilinearity means, e.g., that the decomposition is linear in $\mathbf{a}_s$, when
 $\mathbf{a}_o$ and $\mathbf{a}_p$ are fixed.} We generalize this concept to general nonlinear maps. In particular,
we
define a model for a third-order tensor with index embeddings in the following way:
\begin{mydef}
A model for a third-order tensor with index embeddings is a function
$(\mathbf{a}_s, \mathbf{a}_p, \mathbf{a}_o) \mapsto f(\mathbf{a}_s, \mathbf{a}_p, \mathbf{a}_o)$, where
$\mathbf{a}_s, \mathbf{a}_p, \mathbf{a}_o$ are vectors of real numbers and are called factors, representations, or embeddings for the entity $s$, predicate $p$, and entity $o$, respectively.
\end{mydef}
A generalization to tensors with an order other than three is straightforward.
 We consider tensor models, where, e.g., a neural network might be used to model $\textit{f}^{\textit{prob}}_{\mathbf{w}}(\cdot)$. In that case, $\mathbf{w}$ would be the parameters in the neural network.

As an example, we can obtain a model for the probabilistic KG as
$P(y_{s, p, o, t} = 1) = \textrm{sig} ( \textit{f}_{\mathbf{w}}(\mathbf{a}_s, \mathbf{a}_p, \mathbf{a}_o) )$,
where $\textrm{sig}$ is the logistic function. In this paper, we do not employ tensor models for KGs directly: Instead, we derive tensor models for describing conditional probabilities in stochastic serializations of KGs, as described next.

\section{Stochastic Serialization of Knowledge Graphs and Perception}
\label{sec:serial}

In the following, we are concerned with an agent's knowledge and belief,  instead of ground truth facts. Thus, the Boolean $\textit{KG}_{static}$, $\textit{KG}_{temp}$ and the probabilistic $\textit{KG}_{prob}$,  directly relate
to an agent's belief about facts in the real world, i.e.,  to triples that an agent knows or believes to be   either true or false.  We propose that human memory recall is
an active process by which triples are generated in a sampling process:
Thus it is not sufficient to have a memory, the memory needs to communicate content. In Section~\ref{sec:disc1} we are also concerned with the inverse process: when and how do generated triples refer to statements,  the agent considers true.

Next,  we can serialize the knowledge graphs and derive triple generating processes.

\subsection{A Triple Generating Process for the Static and the Probabilistic Knowledge Graph}

For $\textit{KG}_{\textit{static}}$ we define a categorical distribution
\begin{equation}\label{eq:sampsem}
\mathbb{P}(S=s, P=p, O=o) = \frac{y_{s, p, o}}{\sum_{s, p, o} y_{s, p, o}} .
\end{equation}
By sampling from this distribution, we randomly generate true triples.
By the last equation, we have transformed a knowledge graph into a probabilistic model! We can also consider the reverse direction: If the sampling process generates a triple, we can consider it to be an element of $\textit{KG}_{\textit{static}}$.  Similarly, we can derive a $\mathbb{P}(S=s, P=p, O=o, T=t)$ for the temporal KG.

We can also define
\begin{equation}\label{eq:sampsemapp}
\mathbb{P}(S=s, P=p, O=o) = \frac{\gamma_{s, p, o}}{\sum_{s, p, o} \gamma_{s, p, o}}
\end{equation}
and we could generate triples  with a high likelihood  under the probabilistic KG.

\subsection{Forward Sampling}

Using the chain rule (a.k.a. the general product rule), we can decompose
\[
\mathbb{P}(S=s, P=p, O=o) =
\mathbb{P}(S=s) \;
\mathbb{P}(O=o | S=s) \;
\mathbb{P}(P=p | S=s, O=o) .
\]
We can now easily generate samples by ancestral sampling, a.k.a. forward sampling: First we generate an $s^*$ from $\mathbb{P}(S)$, then an $o^*$ from $\mathbb{P}(O|S=s^*)$, and then a $p^*$ from $\mathbb{P}(P|S=s^*, O=o^*)$.

Thus the content in the static KG  is represented by a triple-generating process, which produces sequences of triples that are elements of the static  KG! A similar process can be generated for the temporal KG.


\subsection{A Triple Generating Process for Perception}
\label{sec:post}

In this section we only consider predicates with visual groundings, like \textit{nextTo}, \textit{looksAt}, \textit{inFrontOf}. These are relevant for perception.
 At time instance $t$, the agent focuses on three image bounding boxes, which were selected
by an attention mechanism.\footnote{The details on this attention mechanisms are not the topic of this paper. In the experiments, we use faster R-CNN \cite{ren2015faster}.}
 We assume
 a bounding box for a subject $\textit{BB}_{\textit{sub}}$ which contains entity $S$,
 a bounding box for an object $\textit{BB}_{\textit{obj}}$ which contains entity $O$,
and a bounding box for a predicate $\textit{BB}_{\textit{pred}}$ specific to predicate $P$. See Figure~\ref{fig-BB}
(image from \cite{lu2016visual}).
{In perception, we are not as much interested in all triple statements that are true at time instance $t$, but in the ones which the agent can infer to be true, given the observed scene at time instance $t$. Another view would be to infer the set of true triples that are causes for the scene.}

\begin{figure}[t]
\vspace{-1cm}
\begin{center}
\includegraphics[width=0.8\linewidth]{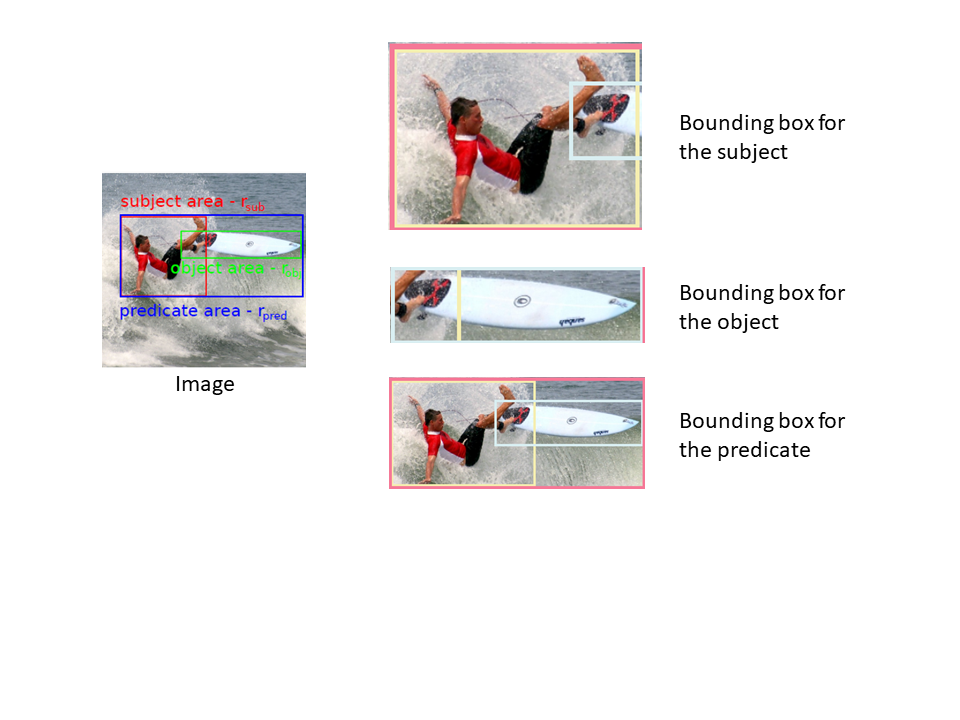}
\end{center}
\vspace{-2.5cm}
\caption{Image and a triple of bounding boxes. Extracted triples could be
\textit{(Person, nextTo, Surfboard)} and \textit{(Jack, nextTo, Surfboard)}.
}
\label{fig-BB}
\end{figure}

\begin{mydef}
A triple-generating process for perception is defined by the conditional probabilities\footnote{Among natural languages with a word order preference, \textit{(s, o, p)} is the most common type \cite{crystal:1997}. This order works best for us since $s$ and $o$ can be predicted with some certainty in most images, whereas $p$ is more of a challenge. }
\begin{equation}\label{eq:subject}
\mathbb{P}(S=s| \textit{BB}_{\textit{sub}}, \textit{BB}_{\textit{pred}}, \textit{BB}_{\textit{obj}})
\end{equation}
\begin{equation}\label{eq:object}
\mathbb{P}(O=o| S=s, \textit{BB}_{\textit{sub}}, \textit{BB}_{\textit{pred}}, \textit{BB}_{\textit{obj}})
\end{equation}
\begin{equation}\label{eq:predicate}
\mathbb{P}(P=p| O=o, S=s, \textit{BB}_{\textit{sub}}, \textit{BB}_{\textit{pred}}, \textit{BB}_{\textit{obj}}) .
\end{equation}
\end{mydef}
As discussed before, this decomposition permits forward sampling, and, if the perception system has low uncertainty, the sampled triples can  conceptionally be  added to the temporal KG.

\subsection{Tensor Models with Index Embeddings}

 We derive tensor models with index embeddings directly for the probabilistic decompositions.

\subsubsection*{Perception's Triple Generating Processes}

A tensor model involved in perception (Equation~\ref{eq:predicate}) is
\begin{equation}\label{eq:perception}
 \mathbb{P}(P =p| S=s, O=o, \textit{BB}_{\textit{sub}, t}, \textit{BB}_{\textit{pred}, t}, \textit{BB}_{\textit{obj}, t})
 = \frac{\exp f^{pred}_{\mathbf{w}}(\mathbf{a}_s, \mathbf{a}_p, \mathbf{a}_o, \mathbf{a}_t)}{\sum_p \exp f^{pred}_{\mathbf{w}}(\mathbf{a}_s, \mathbf{a}_p, \mathbf{a}_o, \mathbf{a}_t)}
 \end{equation}
 \[
\equiv \textrm{softmax}_P (f^{pred}_{\mathbf{w}}(\mathbf{a}_s, \mathbf{a}_p, \mathbf{a}_o, \mathbf{a}_t))
\]
where $f^{pred}_{\mathbf{w}}(\mathbf{a}_s, \mathbf{a}_p, \mathbf{a}_o, \mathbf{a}_t)$ is a fourth-order tensor model and where $\mathbf{a}_t$ is the latent representation for time instance $t$.

For our purposes, we assume
that
 $\mathbf{a}_t$ is a concatenation of representations derived from the bounding boxes,
 \begin{equation}\label{eq:visual}
 \mathbf{a}_t = [\mathbf{a}(\textit{BB}_{\textit{sub}, t}); \mathbf{a}(\textit{BB}_{\textit{pred}, t}); \mathbf{a}(\textit{BB}_{\textit{obj}, t}) ] .
 \end{equation}
Note that the term embedding is preferably used for latent vectors that are static and are tied to concepts (and later are described by connection weights), whereas the term latent representations is preferably used for latent vectors that are, e.g., calculated as functions of sensory inputs.

We share embeddings and latent representations in the conditional probabilities and define
\begin{eqnarray}
 \mathbb{P}(S =s | \textit{BB}_{\textit{sub}, t}, \textit{BB}_{\textit{pred}, t}, \textit{BB}_{\textit{obj}, t}) &=& \textrm{softmax}_S (\textit{f}^{\textit{sub}}_{\mathbf{w}}(\mathbf{a}_s, \mathbf{a}_t)) \label{eq:ssi} \\
 \mathbb{P}(O =o| S=s,\textit{BB}_{\textit{sub}, t}, \textit{BB}_{\textit{pred}, t}, \textit{BB}_{\textit{obj}, t}) &=& \textrm{softmax}_O (\textit{f}^{\textit{obj}}_{\mathbf{w}}(\mathbf{a}_s, \mathbf{a}_o, \mathbf{a}_t)) \label{eq:soi} \\
 \mathbb{P}(P =p| S=s, O=o,\textit{BB}_{\textit{sub}, t}, \textit{BB}_{\textit{pred}, t}, \textit{BB}_{\textit{obj}, t}) &=& \textrm{softmax}_P (\textit{f}^{\textit{pred}}_{\mathbf{w}}(\mathbf{a}_s, \mathbf{a}_p, \mathbf{a}_o, \mathbf{a}_t)) .
 \label{eq:spi}
\end{eqnarray}
In our proposed model, the equations are implemented in form of a special recurrent neural network. See Section~\ref{sec:bioImp} and Figure~\ref{fig-Architecture} in the appendix.


%
%

\subsubsection*{Triple Sampling Process for the Probabilistic KG}

We now propose to use the same functions with the same $\mathbf{w}$, i.e.,
$\textit{f}^{\textit{sub}}_{\mathbf{w}}(\cdot)$,
$\textit{f}^{\textit{obj}}_{\mathbf{w}}(\cdot)$, and
$\textit{f}^{\textit{pred}}_{\mathbf{w}}(\cdot)$,
 for sampling from the probabilistic KG.
Firstly, model sharing is a good idea in general, since it can lead to better generalization,
and secondly, from a biological point of view, it would be wasteful to have separate models for semantic memory and for perception.

We obtain,
\begin{eqnarray}
 \mathbb{P}(S =s) &=& \textrm{softmax}_S (\textit{f}^{\textit{sub}}_{\mathbf{w}}(\mathbf{a}_s, \mathbf{\bar a})) \label{eq:ss} \\
 \mathbb{P}(O =o| S=s) &=& \textrm{softmax}_O (\textit{f}^{\textit{obj}}_{\mathbf{w}}(\mathbf{a}_s, \mathbf{a}_o, \mathbf{\bar a})) \label{eq:so} \\
 \mathbb{P}(P =p| S=s, O=o) &=& \textrm{softmax}_P (\textit{f}^{\textit{pred}}_{\mathbf{w}}(\mathbf{a}_s, \mathbf{a}_p, \mathbf{a}_o, \mathbf{\bar a})) . \label{eq:sp}
\end{eqnarray}
The first line is the conditional probability for the subject, the second one for the object, given the subject, and the last one for the predicate, given the subject and the object.

In the transition from Equations ~\ref{eq:ssi}, \ref{eq:soi}, \ref{eq:spi} to Equations~\ref{eq:ss}, \ref{eq:so}, \ref{eq:sp}
 we
 substituted
\begin{equation} \label{eq:subs}
\mathbf{a}_t \rightarrow
 \mathbf{\bar a}
 \end{equation}
 where $\mathbf{\bar a}$ is the time-invariant probabilistic KG embedding vector, learned from data.
 In a biological interpretation $\mathbf{\bar a}$ becomes the semantic memory embedding (see Section~\ref{sec:discussion}).
 The underlying (reasonable) assumption is that
both the probabilistic model and the perception model can use the same basis,
thus in this sense, the probabilistic KG model (i.e., the semantic memory) is a conjugate prior distribution for the perception model.

The triple sampling process for the probabilistic KG is related to the perceptual process. Thus  the underlying $\gamma_{s, p, o}$  would the  proportion of times that $(s, p, o)$ can be extracted from the image and not
the proportion of times that $(s, p, o)$  is true, as defined before.

\section{Implementation Details}
\label{sec:disc1}

\subsection{Perception Beyond Entities}

So far, we considered graph nodes to be entities. Now we extend the model such that a node can stand for a general concept, e.g., an entity, a class, an attribute, or a location (Figure~\ref{fig-KG}E). Thus the bounding box, e.g., for the subject, only contains one thing, but this thing can have different labels, like \textit{Jack} or
\textit{Person} or
\textit{Happy. }

Sampled triple statements from a fixed triple of bounding boxes, e.g., might now be
\textit{
(Jack, looksAt, Mary),
(Person, looksAt, Mary),
(Jack, looksAt, Person),
(Person, looksAt, Person),
(Happy, looksAt, Mary),
(Happy, looksAt, Blond)}, and so on.

{Recall, that we associate a time instance $t$ with a bounding box triple.} This is important such that, e.g., the triple
\textit{(Happy, looksAt, Mary)} is meaningful and can be associated with the correct bounding boxes.

%


\subsection{Chatterbox Decoding}

Clearly, sampling might also produce semantically contradicting sentences.
As an example, if for the same three bounding boxes the model generates the triples
\textit{(Jack, looksAt, Mary)}, \textit{(Person, looksAt, Mary)} and \textit{(Dog, looksAt, Mary)},
 then the agent would not catch this semantic contradiction.
 It would simply conclude that Jack has properties ``person'' and ``dog''.
Our approach does not check for semantic conflicts
on the declarative sentence level. We continue this discussion in more detail in Section~\ref{sec:discussion}.

%

\subsection{Properties and a Simple Semantic Decoder}

If $s_1^*$ and $s_2^*$ are two samples for the same bounding box, the agent concludes
that \textit{($s_1^*$, hasProperty, $s_2^*$)} is true. For example, if \textit{Jack} is a label for a bounding box, and \textit{Happy} is a label for the same bounding box, then the agent concludes that
\textit{(Jack, hasProperty, Happy) }is true and, conceptionally, the fact is entered in $\textit{KG}_{\textit{temp}}$. Other examples are,
\textit{(Jack, hasProperty, Person)},
\textit{(Jack, hasProperty, Happy)},
\textit{(Mary, hasProperty, Blond)}.
As a postprocessing step, \textit{hasProperty} can be mapped to more meaningful predicates with a representation in the schema. Thus, e.g., \textit{(Mary, hasProperty, Blond)} becomes \textit{(Mary, hasHaircolor, Blond)} and \textit{(Jack, hasProperty, Person)} becomes \textit{(Jack, hasClass, Person)}.

For this decoding of properties, a simple decoder is sufficient.
The agent only needs to repeatedly sample from the same bounding box, let's say the subject, using Equation~\ref{eq:subject}, and generate several bounding box labels.
We discuss simple decoding in more detail in Section~\ref{sec:discussion}.

\subsection{From a Bag of Samples to a Set of Facts for the Temporal KG}
\label{sec:cleanup}

Assume that, in ground truth, a bounding box triple is described by $N_s$ true
 sentences. If we generate a bag of $N>> N_s$ samples $\{s_i^*, p_i^*, o_i^*\}_{i=1}^N$, using Equations~\ref{eq:ssi},~\ref{eq:soi} and~\ref{eq:spi}, and remove
 multiple instances of the same samples, we recover
 a subset of the true triples.
 There is a trade-off: If $N$ is selected too small, we might miss some of the true triples and if we select $N$ to be too large, we also produce erroneous triples.
Generated triple sets will be used to train the perceptual system and the probabilistic KG, as described next.

\subsection{Machine Learning for Perception}
\label{sec:train}

\subsubsection*{Supervised Learning for Perception}

The training data targets are represented in the adjacency tensor for the temporal knowledge graph.
Assume that at time instance $t^*$, a true observed triple is
$(s^*, o^*, p^*)$.
 Its contribution to the cross entropy cost function for the perceptual model is then
\[
\textit{cost}_{\textit{perc}, s^*, p^*, o^*, t^*} =
\]
\[
- \log \textrm{softmax}_S (\textit{f}^{\textit{sub}}_{\mathbf{w}}(\mathbf{a}_{s^*}, \mathbf{a}_{t^*}))
-\log \textrm{softmax}_O (\textit{f}^{\textit{obj}}_{\mathbf{w}}(\mathbf{a}_{s^*}, \mathbf{a}_{o^*}, \mathbf{a}_{t^*}))
- \log \textrm{softmax}_P (\textit{f}^{\textit{pred}}_{\mathbf{w}}(\mathbf{a}_{s^*}, \mathbf{a}_{p^*}, \mathbf{a}_{o^*}, \mathbf{a}_{t^*}))
\]
with the decomposition in Equations~\ref{eq:ssi},~\ref{eq:soi}, and~\ref{eq:spi}.

\subsubsection*{Self-supervised Learning for Perception}

In self-supervised learning, without labeled examples, we use exactly the same cost function,
only that $(s^*, o^*, p^*)$ is a sample generated in the perception sampling using
 Equations~\ref{eq:ssi},~\ref{eq:soi} and~\ref{eq:spi} and with the removal of multiple copies, as described in Section~\ref{sec:cleanup}.
%
%
 This form of self-supervised learning can be considered a form of Monte Carlo EM \cite{wei1990monte}, where the Monte Carlo E-step
corresponds to the generation of the samples, and the maximization step to the gradient update, taking the samples as actual data.

\subsection{Machine Learning for the Probabilistic KG}
\label{sec:trainsm}

We now consider the adaptation of the probabilistic KG embedding vector $\mathbf{\bar a}$ (introduced in Equation~\ref{eq:subs}).
If $(s^*, p^*, o^*)$ is a training triple at time instance $t^*$,
\[
\textit{cost}_{\textit{prob}, s^*, p^*, o^*, t^*} =
\]
\[
- \log \textrm{softmax}_S (\textit{f}^{\textit{sub}}_{\mathbf{w}}(\mathbf{a}_{s^*}, \mathbf{\bar a}))
-\log \textrm{softmax}_O (\textit{f}^{\textit{obj}}_{\mathbf{w}}(\mathbf{a}_{s^*}, \mathbf{a}_{o^*}, \mathbf{\bar a}))
- \log \textrm{softmax}_P (\textit{f}^{\textit{pred}}_{\mathbf{w}}(\mathbf{a}_{s^*}, \mathbf{a}_{p^*}, \mathbf{a}_{o^*}, \mathbf{\bar a}))
\]
with the decomposition in Equations~\ref{eq:ss}, ~\ref{eq:so}, and ~\ref{eq:sp}.

 For visual triples, the learned probabilities are all w.r.t. the perceptual process. E.g., for semantic memory, the conditional probabilities are prior probabilities:
$\mathbb{P}(S=\textit{John})$ is the prior probability that
the next sample generated from the subject bounding box refers to \textit{John},
$\mathbb{P}(O = \textit{Mary} | S=\textit{John})$ is the prior probability that the next sample generated from the object bounding box refers to \textit{Mary} if the subject bounding box sample was \textit{John}, and
$\mathbb{P}(P = \textit{looksAt} | S=\textit{John}, O = \textit{Mary})$ is the prior probability that the next sample generated from the predicate bounding box refers to \textit{looksAt}
if the subject bounding box sample was \textit{John} and the object bounding box sample was \textit{Mary}.

In operation, we can train the probabilistic KG in a self-supervised manner with the samples generated from perception, as well.


Self-supervised learning can also be performed using triples generated by the probabilistic KG.
In particular, this is important for the triples in the probabilistic KG that are shared with the static KG, typically related to background information. For example, \textit{(Jack, looksAt, Mary)} might be a visual triple statement and $(Jack, sibling, Mary)$ is a background triple statement, which might not have been acquired by visual perception. Since these triples are ''true forever`` they should be sampled regularly in self-supervised training, and used in the
 adaptation of the probabilistic KG, to avoid forgetting.\footnote{In the experiments, the background probabilistic KG
is a separate categorical model with its own   $\mathbf{\bar a}^{bg}$.}

Some predicates might have two representations: E.g., ``hairColor'' might appear as
``hairColorObserved'' in perception and ``hairColorTrue'' in background.


\section{Towards a Cognitive Architecture}
\label{sec:bioImp}

\subsection{From KGs and Tensors to Perception and Semantic Memory}

We now begin relating the technical model to human perception and memory.
First, we propose that the triple generation process in our perception model
 can be related to human perception.
Second, we propose that the triple generation process for our probabilistic KG can be related to human semantic memory.

\subsection{Biological Constraints}

We now consider implementations of the developed equations for perceptual decoding.
Standard tensor models with index embeddings
 cannot easily be implemented in brainware.
First, most tensor factorizations, such as the canonical polyadic decomposition, the Tucker decomposition and RESCAL, require the excessive multiplication of factors; multiplication is an operation that is not easily implemented in biological hardware.
Second, tensor models are functions of several embedding vectors and, following our single brain hypothesis (discussed in Section~\ref{sec:discussion}), only one thing at a time can be represented in the brain's representation layer (see Figure~\ref{fig-arch}). This results in the need for an extra memory layer, i.e., the working memory.
Third, for processing speed, we propose that there needs to be a direct path from sensory input to concept prediction, without a detour through a memory layer. The contribution of the memory layer to concept prediction should be a modification of this direct prediction path.
Fourth, we also need to take into account the sequential nature of the processing, so processing must proceed forward in time.
A detailed analysis of our model in the context of current discussions in cognition and neuroscience follows in Section~\ref{sec:discussion}.

\subsection{Overall Architecture}

We propose that a minimal implementation of a perceptual sampling process, following
Equations~\ref{eq:ssi}, \ref{eq:soi}, and \ref{eq:spi}, under the discussed biologically motivated constraints,
 requires the layers shown in Figure~\ref{fig-arch}.
The central communication platform is the representation layer $\mathbf{q}$.
It receives information about the visual input provided by the sensory memory layer $\mathbf{g}$.
The index layer $\mathbf{e}$ contains the indices for concepts.
The representation layer can activate the indices and vice versa.
 To be able to decode triples, and for memory functions in general, we introduce the working memory layer $\mathbf{h}$.
 It receives input from the representation layer and also transmits information back to the representation layer.
 Layer $\mathbf{h}$ has self-recurrent connections.

\subsection{Processing Steps}

We first consider the decoding of the subject.
The simplest implementation of Equation~\ref{eq:ssi} is as follows. Let $\mathbf{g}$
describe the output of the visual processing system. We implement the visual processing system by a deep neural network mapping where the input is the subject bounding box $\textit{BB}_{\textit{sub}}$. By connection matrix $D$, this information activates the representation layer $\mathbf{q}$, which then activates layer $\mathbf{e}$ via connection weights $A^T$.
The $i$-th column of matrix $A$ consists of the embedding vector $\mathbf{a}_{e_i}$.

Now we sample from the ensuing categorical distribution (Equation~\ref{eq:ssi}) and the brain makes a commitment to a unique interpretation, by selecting a sample $s^*$.
The selected
$s^*$ activates the representation layer as $\mathbf{q}:= \mathbf{a}_{s^*}$, and, by this activation of the representation layer, the complete brain is informed about the selected entity.

We essentially repeat the procedure also for the object, Equation~\ref{eq:soi}, and the predicate, Equation~\ref{eq:spi}. But now we need to involve the working memory layer $\mathbf{h}$. This layer stores anything that has been going on in the representation layer so far and it contributes this information back to the representation layer.

Thus for the object, the representation layer is activated by visual input from $\textit{BB}_{\textit{obj}}$
via the deep neural network, and from the working memory layer $\mathbf{h}$ via connection matrix $W$. Again, the representation layer
 then activates layer $\mathbf{e}$ via connection weights $A^T$.
 We sample an $o^*$ and we then set, $\mathbf{q}:= \mathbf{a}_{o^*}$.

We proceed similarly for the predicate, only that the visual input now is coming from $\textit{BB}_{\textit{pred}}$.
We sample a $p^*$ and we then set, $\mathbf{q}:= \mathbf{a}_{p^*}$.

After decoding an $(s^*, p^*, o^*)$-sample,
we can do another pass through the working memory,
 which integrates information about the triple and can also serve as initialization for follow-up triple decoding, providing context.
 The representation layer finally is a representation of the complete decoded triple sentence.

A detailed description of a
concrete implementation can be found in the appendix. See, in particular, Figure~\ref{fig-Architecture}.
Although the architecture is related to the decoder in encoder-decoder architectures used in machine translation \cite{cho2014learning}, there are also significant differences, mainly due to the fact that we insist on a direct path from sensory input to concept prediction.
Figure~\ref{fig-arch} also includes indices for time instances, which are important for episodic memory and an index $\bar e$ for episodic memory.
Sampling $\bar e$ would activate the representation layer with
memory embedding vector $\mathbf{\bar a}$,
and the model produces samples from the semantic memory.
See our discussion in Section~\ref{sec:discussion} and Figure~\ref{fig-Architecture} in the appendix.

\begin{figure}[t]
\vspace{-1cm}
\begin{center}
\includegraphics[width=0.8\linewidth]{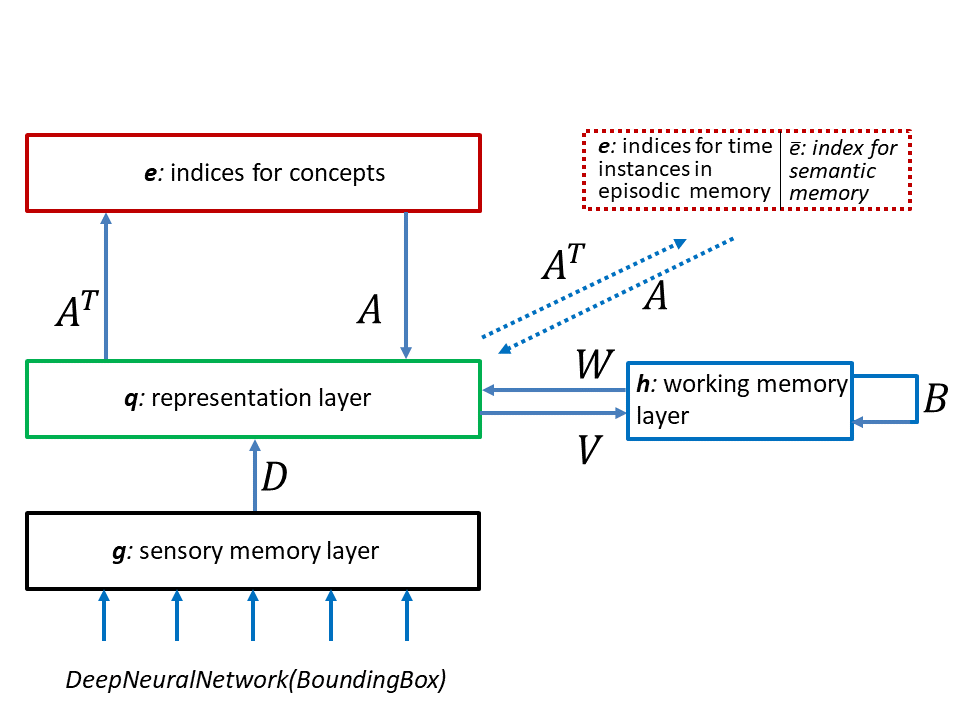}
\end{center}
\vspace{-0.5cm}
\caption{Our model architecture consists of four layers. Extracted representations from images are represented at the bottom layer activating the sensory memory,
$\mathbf{g}$. This layer is connected to the representation layer $\mathbf{q}$.
The top layer $\mathbf{e}$ contains the indices for entities and predicates.
The working memory $\mathbf{h}$ is a short term memory.
The dotted segment includes the time indices $e_t$ of episodic memory and the semantic memory index $\bar e$ for semantic memory. See Section~\ref{sec:discussion}.}
\label{fig-arch}
\end{figure}

\section{Relationships to Information Processing in the Brain}
\label{sec:discussion}

In this section, we relate our model to current discussions in cognition and neuroscience. Main points are formulated as propositions.
A speculative assignment of the model layers to brain areas is shown in Figure~\ref{fig-archBrain}.

\begin{figure}[t]s
\vspace{-1cm}
\begin{center}
\includegraphics[width=0.8\linewidth]{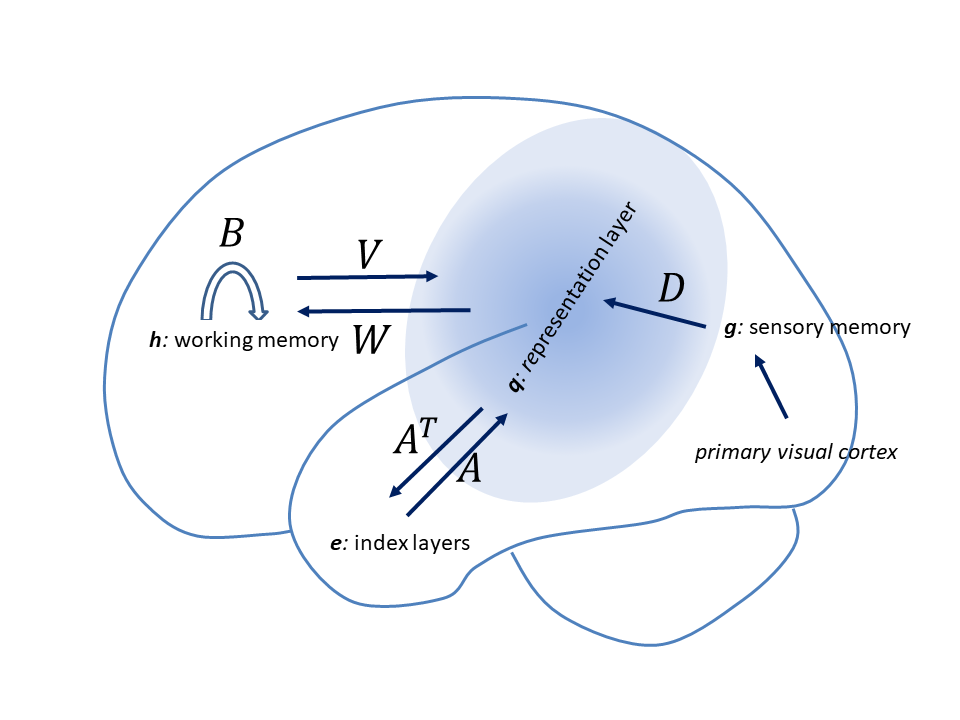}
\end{center}
\vspace{-0.5cm}
\caption{There is still great uncertainty about the localization of the different functions in the brain. Here is one plausible allocation.
The model's layer $\mathbf{g}$ would correspond to the visuospatial sketchpad in the parietal-occipital region.
The model's index layer $\mathbf{e}$ would correspond to indices formed in the medial temporal lobe (MLT).
Indices for concepts are consolidated in hubs like the
anterior temporal lobe (ATL). Some papers also see a greater role of the posterior
parietal cortex, as part of a parietal memory network (PMN).
 MTL and other structures, like the medial prefrontal
cortex (mPFC), are involved in the consolidation of episodic memories. 
Our working memory $\mathbf{h}$ is part of the brain's working memory
which involves the
prefrontal parietal network (PPN). 
The brain's version of the representation layer $\mathbf{q}$ is distributed across the neocortex, including sensory and motor centers.
}
\label{fig-archBrain}
\end{figure}

\subsection{Sensory Memory Layer and Sensor Encoding}

In our model, we use a deep convolutional neural network for the mapping from the bounding boxes to the visual sensory memory.
The mapping of deep convolutional neural networks to the operations of different brain regions ---not discussed in this paper--- is a focus of current research, e.g., \cite{kriegeskorte2018cognitive}.

Layer $\mathbf{\mathbf{g}}$ in our model would be related to the visual sensory memory, maintaining visual information to be processed and analyzed, and it represents properties of the respective visual focus of attention.
In the brain,
sensory memory (time scale of milliseconds to a second) represents the ability to retain impressions of sensory information after the original stimuli
have ended \cite{tulving1972episodic,coltheart1980iconic,gazzaniga2004cognitive}.
It is assumed that the visual sensory memory layer involves
the visuospatial sketchpad of the working memory, associated with the parietal-occipital region of the brain \cite{gazzaniga2004cognitive}.

\subsection{Concept Engrams}
\label{sec:index}

An engram is a memory trace in the brain. In this section, we focus on engrams for concepts such as entities,
classes, locations,  and predicates, representing conceptual knowledge \cite{ralph2017neural}.
Engrams for episodic memory are discussed further down.

\begin{myhyp}
In the brain, concept engrams are memory traces, e.g.,
for entities, classes, locations, attributes,  and predicates.
In our model, an engram
 consists of a concept index $e$, realized by a unit in the index layer $\mathbf{e}$,
 and its concept embedding, which is
 realized as a connection vector $\mathbf{a}_e$
 connecting an entity index $e$ with the representation layer $\mathbf{q}$.
 \end{myhyp}
\begin{myhyp}
 Since embeddings are optimized for their different roles in perception and memory,
 they implicitly reflect all that is known about a concept. In the brain, embeddings would be a part of an implicit concept memory, but they are also required for semantic decoding.
\end{myhyp}
The concept index by itself has an amodal, local, symbolic character, whereas the embedding has a modal, distributed character, and grounds the concept.

The debate about localized representations in the brain is ongoing; currently, there are many experimental results pointing towards locality \cite{kiefer2012conceptual}.
As ~\cite{dehaene2014consciousness} puts it: ``Every cortical site holds its own specialized piece of knowledge''.
 In agreement with current discussions,
we are favoring localized representations of concepts in the biological equivalent of the index layer $\mathbf{e}$ and localized semantic interpretations
 in the representation layer $\mathbf{q}$: 
 \cite{binder2011neurobiology} states: ``The neural systems specialized for storage and retrieval of
semantic knowledge are widespread and occupy a large
proportion of the cortex in the human brain.''

 Naturally, locality of representation is probably only discovered in well-designed experiments:
In our model, an activated concept index activates several units in the representation layer, and a unit in the representation layer, in turn, activates many indices. Since index activations might change rapidly, the general appearance might be that of a globally activated system, hiding locality of representation.

Specific concept cells have been found in the medial temporal lobe (MTL) region of the brain. MTL includes the hippocampus, along with the
surrounding hippocampal region consisting of the perirhinal, parahippocampal, and entorhinal neocortical regions.
In particular, researchers have reported on a remarkable subset of MTL neurons
that are selectively activated by strikingly different pictures of given individuals, landmarks or objects,
 and in some cases even by letter strings with their names \cite{quiroga2012concept,quiroga2005invariant}.

We argue that the components or units in the biological realization of the representation layer $\mathbf{q}$ have specific meanings as well. 
 In different brain regions, maps have been discovered that code for visual appearance, sound, and function. These regions would be the units in the biological equivalent of our representation layer $\mathbf{q}$ and could be activated by the concept indices.
For instance, the concept “cat” includes
the information that a cat has four legs, is furry, meows, can
move or can be petted \cite{kiefer2012conceptual}.
As another example, consider the recall of the concept ``hammer'', represented in the index layer $\mathbf{e}$ \cite{rueschemeyer2010function}. This might excite brain
areas indicating a typical hammer appearance, the sound of hammering and the required motor movement of hammering,
 all represented in the biological equivalence of representation layer $\mathbf{q}$. In our model, since $\mathbf{q}$ is activated by visual input, the embeddings will mostly be visually grounded, as well.

Evidence for distributed semantic activation has, e.g., been described by \cite{Huth2016}.
That paper developed a detailed atlas of semantic categories and their topographic organization by extensive fMRI studies, showing the involvement of
the lateral temporal cortex, the ventral temporal cortex,
the lateral parietal cortex, the medial parietal cortex,
the medial prefrontal cortex, the superior prefrontal cortex,  and the inferior prefrontal cortex \cite{Huth2016}.

\cite{ralph2017neural} has defined a hub-and-spoke model.
The hub is supposed to be located in the anterior temporal lobes (ATLs), which might be where concept indices
 are consolidated after
they are formed in MTL. The hub is connected to several different areas (e.g., visual cortex, auditory cortex, orbitofrontal cortex), which might be part of the biological realization of the representation layer.
Other hubs have been located in the frontal, the temporal and the parietal lobe \cite{tomasello2017brain}.

For spatial memory, the posterior parietal cortex fulfills all criteria for a
hippocampus-independent memory representation \cite{brodt2016rapid}. Researchers also discuss a parietal memory network (PMN) \cite{gilmore2015parietal}.

Recently, evidence has been found that embeddings are context dependent \cite{popp2019processing}. In our model, the representation layer is activated by sensory input
and by context indices,
so the model is informed about a concept in context, even with the connection weights between layers $\mathbf{e}$ and $\mathbf{q}$ being fixed. The feedback from the working memory layer also represents context which affects representations.
Other mechanisms for realizing context-dependent representations in the brain could be attention based
in combination with context-specific concept
 indices.

In our model, we have a bipartite relationship between the concept index layer $\mathbf{e}$ and the representation layer $\mathbf{q}$.
The cortical network, in general, is not strictly bipartite and contains extensive local connectivity, realizing an interplay
of synaptic excitation and synaptic inhibition \cite{isaacson2011inhibition}.
 Thus bipartiteness might not be true in the actual brain,
but it might be a reasonable approximation to biological reality.
A specific entity has very specific embeddings, so strong links between an entity index and associated nodes in the representation layer are quite plausible.
Considering the diversity of concepts, strong correlations among representation nodes might be rarer and learning
 might favor independent representations~\cite{hyvarinen1999fast}.
Dependencies between concepts in our model are mediated via activation of the representation layer. We would not exclude direct entity-entity links, but again, they might be rare.
The advantage of a purely bipartite architecture is interpretability and speed of operation. Considering that in our model $\mathbf{q}$ was activated by visual input, the activation of the concept indices then is simply the
last layer in a classifier.
Without a bipartite structure,
decoding is an iterative process requiring convergence to an equilibrium point, potentially
using stochastic sampling, as in the restricted Boltzmann machine \cite{smolensky1986information,hinton2010practical}.

In the brain one typically encounters sparse distributed representations \cite{rolls2016cerebral}.
Sparsity in the embedding vectors can be achieved in technical models, e.g.,
 by using appropriate regularization terms, like LASSO \cite{tibshirani1996regression}.
 Sparsity can also be encouraged by enforcing nonnegativity of model parameters.

In our model, the weights between $\mathbf{e}$ and $\mathbf{q}$ are symmetrical.
The biological plausibility of symmetric weights has been discussed intensely in computational neuroscience and many biologically oriented models
have that property.
There is a significant amount of evidence that neural systems are bidirectional in the sense that if a set of neurons is connected with another set of neurons,
connections in the back-direction also exist. But it is generally not assumed that connections are symmetrical, rather back-projecting connections are
typically less strong and numerous than forward connections \cite{rolls2016cerebral}.
In our model, we can relax the symmetry requirement: Symmetry makes sense, but it is not essential, since we are not dependent on some fixed point stability, as other models \cite{hopfield1982neural,hinton2006reducing}.

\subsection{Cognitive Maps and Schemas}

Cognitive maps are well established in spatial domains, as grid cells in the mammalian entorhinal cortex \cite{rowland2016ten,stachenfeld2014design,constantinescu2016organizing}.
They encode the geometry of allocentric space
with a periodic ‘‘grid’’ code \cite{hassabis2017neuroscience}.
This grid-like code might also be present in cortical regions as the medial frontal, the medial parietal and lateral temporal cortices, and possibly even for nonspatial abstract conceptual representations \cite{constantinescu2016organizing}.
 Spatial memories, as other memories, are thought to
be slowly induced in the neocortex by a gradual
recruitment of neocortical memory circuits in
long-term storage of hippocampal memories \cite{mcclelland1995there,squire1995retrograde,frankland2001alpha,moser2015place}.

Representations for concepts
form what is called a \textit{schema}. \cite{moscovitch2016episodic} defines a schema as ``adaptable associative networks of knowledge extracted over multiple
similar experiences''.
 Studies have shown that individuals
 can analyze perceptual information significantly more easily when
this information is related to an acquired schema.

In our model, an improvement in the concept schema would go hand in hand with a
refined perception, whereas learning about new concepts would require considerable effort.
Forming a new concept means introducing a new concept index and learning an embedding for the new concept. This is related to few-shot learning in technical systems \cite{snell2017prototypical}, but with unlabeled data, which is a challenging problem.

\begin{myhyp}
In our model, dealing with a new concept is much more involved than dealing with a concept that already has an index and an embedding. In the first case, the model needs to realize that there is a new concept, a new index needs to be formed, and a new embedding needs to be learned.
These three steps might also play a role in the brain when new concepts are formed.
\end{myhyp}
Or, as the German writer Goethe put it: ``We only see what we know''.

The cognitive sciences have a certain focus on concept structures. There is considerable work
 on semantic networks \cite{collins1975spreading,collins2004retrieval}, word net \cite{miller1998wordnet}
and formal domain ontologies \cite{staab2010handbook}.
Topic models and mind maps are among the more modern approaches.

In our approach, links between concepts are not stored explicitly but are calculated on-demand from node embeddings, and information propagates via the representation layer. In our model, there are several ways of analyzing concept structure.
First, following our previous discussion, the concept representations projected to the representation layer $\mathbf{q}$
are visually grounded and should be interpretable.
Second, similarity between concepts can be analyzed by clustering in embedding space, as pursued by \cite{nickel2015}.
Third, there are special embedding learning approaches that focus on hierarchical concept structures in embedding models \cite{nickel2017poincare}.
 Fourth, similarity between triple statements can be found by an analysis of the integrated triple representations. This might be useful for the discovery of triple correlations and causal dependencies.

\subsection{Representation Layer}

\begin{myhyp}
The representation layer would correspond to the brain's main ``communication platform'', ``communication bus``, ``blackboard'', or the shared ``canvas'', where it enables a global information exchange.
To convey information, both in our model and the brain, perception,  and memory need to activate
 this layer.
\end{myhyp}
In our model, the representation layer can be activated by perception, semantic memory and, as we will discuss further down, by episodic memory and communicates with sensory memory layer, index layer,  and working memory.
As discussed,
the biological equivalent of the representation layer involves diverse areas, including the motor and association cortex.

In semantic decoding, the representation layer is periodically activated, which might be reflected in neural signals and could be related to some of the neural oscillations found in the brain.
A candidate is the beta rhythm (13-35 Hz) considered to be related to consciousness, perception, and motor behavior. Also of interest is the gamma wave (25-140 Hz), which is correlated with large scale brain network activity and cognitive phenomena such as working memory, attention, and perceptual grouping.

The clear distinction between representation layer and index layer might be blurred in the brain: a unit in the representation layer might sometimes act as an index and vice versa.

\subsection{Semantic Decoding}
\label{sec:semdec}

The simple semantic decoder was introduced in Section~\ref{sec:disc1}.
\begin{myhyp}
In our model and maybe also in the brain, a simple semantic decoder does not require working memory. It is sufficient for decoding of sentences with unary predicates.
\end{myhyp}
First and foremost, a simple decoder uses the direct path from sensory input to the index layer, so concept classification is fast and does not do a detour through the working memory layer.

The simple decoder can associate several labels originating from the same bounding box. Thus it can conclude that the bounding box represents Jack, and it represents something that is blond, so it must be true that \textit{(Jack, hairColor, Blond)}. Here, the predicate-object pair can be represented by the unary predicate
\textit{hairColorBlond}, since ``Blond'' does not have object characteristics.
 A triple with a binary predicate would be \textit{(Jack, knows, Mary)} which, as a relation,
\textit{knows(Jack, Mary)}, has two arguments, since Mary, as a person, does have object characteristics.
\begin{myhyp}
A complex semantic decoder considers the dependencies between subject, object, and predicate.
In our model and, as we propose also in the brain, the complex semantic decoder requires working memory
and builds upon concept indices and concept embeddings.
\end{myhyp}
As our experimental results will show, the complex decoder ---in combination with the working memory layer--- gives much-improved results for decoding sentences with binary predicates. Thus we propose that
 working memory greatly improves the ability of humans to formulate triple statements representing entity-to-entity relationships.
To what degree animals have working memory functionalities is still a matter of debate. \cite{carruthers2013evolution} concludes that other primates have working memory systems
homologous with humans', but that humans are unique in some of its uses, specifically of inner speech.

\subsection{Working Memory Layer}
\label{sec:wm}

A complex semantic decoder requires a memory function.
\begin{myhyp}
Single brain hypothesis: in our model the representation layer $\mathbf{q}$ is memoryless and can only present one thing at a time;
any information that needs to be stored for later processing requires working memory layer $\mathbf{h}$.
We suggest that also the brain needs to employ the working memory for any short-term storage of information.
\end{myhyp}
Deep neural networks process information in a sequence of layered parallel processing steps. The same type of processing occurs in the complex semantic decoder, which activates a sequence of layers, involving layers $\mathbf{g}, \mathbf{q}, \mathbf{e}, \mathbf{h}$, as part of a recurrent structure
 to produce a sequence of concept indices (see Figures~\ref{fig-arch} and~\ref{fig-Architecture}).
\begin{myhyp}
In the semantic decoder, processing is executed as a sequence of layered parallel steps, activating a sequence of concept indices.
\end{myhyp}
Sequential processing is also a core concept in the theory of a global workspace \cite{baars1997theater, bor2012consciousness, dehaene2014consciousness}:
\cite{koch2014keep} discusses that Dehaene's workspace has
extremely limited capacity (``the central bottleneck'') and that the mind can be conscious of only one or
a few items or events, although these might be quickly varying.
In cognitive neuroscience, the general notion is that multitasking likely is an illusion.
The sequential processing would also contribute to a potential solution of the binding problem \cite{singer2001consciousness}, since the decoding focuses on concepts in a serial fashion and associations between activities in the representation layer and the index layer are well defined.

In general, working memory is associated with decision making and cognitive control \cite{baddeley1992working} and is necessary for keeping
task-relevant information active as well as manipulating that information to accomplish behavioral tasks.
There is an emerging consensus that most working memory tasks recruit a network of the prefrontal cortex (PFC) and parietal areas in the prefrontal parietal network (PPN). 
 PPN activity is consistently reported in both attention and consciousness studies \cite{bor2012consciousness}. The latter publication proposes that
the PPN can be viewed as a ``core correlate'' of consciousness.

\cite{dehaene2014consciousness} defines consciousness as ``global information sharing'' 
where information has entered into a specific storage area that makes it
 available to the rest of the brain. 
The biological equivalent of our representation layer $\mathbf{q}$, together with the working memory layer (in our model, $\mathbf{h}$), might have some properties of the brain's global workspace.

Christof Koch and colleagues argue that the posterior hot zone (PHZ) is the minimal neural substrate essential for conscious perception \cite{koch2016neural}. The PHZ includes sensory cortical areas in the parietal, temporal, and occipital lobes.
The biological equivalent of our representation layer (in our model, $\mathbf{q}$) might have some properties of the PHZ.

Another interesting point is that both theories assume mental states, well delineated from all the other states. Such a process is going on in our sampling approach if one interprets a sample as a decision on an interpretation: ``It's a bird, or a plane, or it's Superman, but not all of them at the same time'' \cite{dehaene2017consciousness}.
\cite{dehaene2014consciousness} talks about a similar process of
``... collapsing all unconscious probabilities into a single conscious sample ...''.
His model assumes a ``winning neural coalition'' whereas our sampling approach is much simpler, but maybe a reasonable approximation.
The importance of sampling is also recognized in \cite{dehaene2014consciousness} in the context of conscious perception.
 For example, the author states that ``... consciousness is a slow sampler''.

In our model, working memory is part of a (non-standard) recurrent neural network. \cite{hassabis2017neuroscience} also discusses the connection between recurrent neural networks and working memory.
Working memory, of course, realizes many more functions and is not limited to storing intermediate processing results in complex semantic decoding.

\subsection{From Triple Statements to Language}

Humans differ from other animals in their ability to express themselves in the form of natural languages.
Human language is the basis for communication but also a means to argue and reason.
The \textit{language of thought hypothesis} is the hypothesis that mental representation has a linguistic structure: thoughts are sentences in the mind. \cite{fodor1975language} describes the nature of thought as possessing ``language-like'' or compositional structure (sometimes referred to as mentalese). In this view, simple concepts combine in systematic ways (akin to the rules of grammar in language) to build thoughts.
From internal sentences, generated by the semantic decoder, there is a direct path to external sentences, i.e., to human communication by natural language.
\begin{myhyp}
Both in our model and the human brain, the triple statements generated in perception and memory recall are a basis for a language of thought and might be related to a form of an ``inner speech''.
\end{myhyp}
Our work complements 
publications where the goal has been to automatically generate  image annotations~\cite{cheng2018survey}.
In our model, working memory is involved in semantic decoding.
Indeed, individuals with aphasia can demonstrate deficits in short-term memory, working memory, attention, and executive function \cite{murray2000assessing}. It is generally assumed that language generation involves working memory but also several other areas, such as Broca's area \cite{hickok2007cortical}.

Consider a set of sentences, describing a complete scene.
If $N_{\textit{scene}}$ is the number of bounding box triples in a scene,
 then, in total, one can represent
$\mathcal{O}((N_E^2 N_P)^{N_{\textit{scene}}})$ different scene descriptions.
We truly achieve a form of a compositional benefit, compared to a one-hot encoding of image descriptions.
True perception only becomes possible by biasing this exponential space to the triple statements and sequences of triple statements actually reflecting the statistics in the real world, as discussed next.

Triple statements can drive communication, argumentation and logical reasoning \cite{richardsonmarkov2006,hildebrandt2020reasoning}.
First, there is a clear advantage to communicate with other agents about the part of the world, which is not perceived here and now. Thus an agent can tell another agent not to leave the hide-out since there is a leopard waiting outside, which is hiding, and which cannot directly be perceived.
Second, scenes can be described with great accuracy, as just discussed.
Thus we can act better since we know more. Third, triple statements and language are prerequisites for argumentation and logical reasoning and for studying advanced mathematics and the sciences.
An interesting question is: what came first?
Did we develop
semantic decoding to develop
language and to
communicate ---and improved perception and logical reasoning were by-products, which enabled us to act better---
or did we develop improved perception first, with language and reasoning as by-products?

\subsection{Consistency of a Set of Triple Statements}

In this paper, we are focussing on dependencies between subject, predicate,  and object in a triple statement.
Triple statements are not mutually independent, either, reducing the exponential space of possibilities to combinations of triple statements, likely occurring in the actual world.

\subsection*{Probabilistic KG}

A strength of our tensor models is that concept embeddings
 are mutually coupled in training, which leads to nonlocal dependencies
 that implement an inductive form of reasoning \cite{nickel2015}.
One might argue that this effect is only partially capturing patterns in the explicit triple statements that are, for example, expressible in first-order logic (FOL).
 The work on Markov logic networks (MLNs) \cite{richardsonmarkov2006} is among the most promising approaches, combining learning with FOL expressibility.
Unfortunately, in most technical applications, the success of logical inference
is quite limited: Typically, these models are brittle and difficult to handle and results are mixed.

Certainly, humans are able to reason on the conceptual level, master argumentation and potentially even logical reasoning, and it remains a challenge to develop technical systems of similar power and robustness.
 Checking for logical inconsistencies and doing human-level argumentation and logical reasoning probably needs a fully developed human intelligence including an advanced working memory.

 There is also no ontological consistency check in most current large-scale KGs. For example, the common RFD ontology standard does not contain a construct for negation and cannot produce ontological contradictions.
We would propose that the mind uses also a weak (shallow) ontology early in processing and syntactic and semantic contradictions are
 resolved at a higher level of reasoning.\footnote{Google's Jamie Taylor (http://videolectures.net/jamie\_taylor/) proclaimed that the Google KG is ``RDF minus minus'',
where RDF is already a week ontology. Ontological reasoning simply introduces too many problems in large scale open KGs.}

\subsection*{Perception}

Consider a variant, where, as an additional input, the model is obtaining input from the complete image, not just from the currently selected bounding box triple.
This introduces dependencies between all extracted triples in an image and across all sets of bounding boxes and, thus, introduces dependencies between extracted triple statements, as well. The chaining of triple decoding also introduces triple dependencies (Figure~\ref{fig-Architecture}).

 On the other hand, \textit{a posteriori} commitments are independent in our approach.
For example, assume that the triple \textit{(Sparky, looksAt, Jack)} has a posterior probability of 80\% and
 \textit{(Dog, looksAt, Jack)} has a posterior probability of 80\%. If the agent commits to
 \textit{(Sparky, looksAt, Jack)} being true,
 our model would not change the probability for \textit{(Dog, looksAt, Jack)}, although logical reasoning
 on the declarative triple-statement level should change the probability to 100\%.

Scene interpretations with dependent triple statements can also be achieved by structured prediction. Classical approaches are hidden Markov models or conditional random fields. Dependent commitments are used in many scene graph models \cite{johnson2015image,yang2018graph}. A very recent approach is \cite{bengio2017consciousness}. It attempts to capture dependencies by
 sparse factor graphs.
 Interestingly, leading approaches to text modeling do not perform structured prediction, since output correlations are modeled within the representation layers \cite{devlin2018bert}. Dependent commitments can also be achieved by ontological/logic postprocessing, which belongs to the more recently acquired faculties of the human brain.

In Section~\ref{sec:disc1} we defined chatterbox decoding, which essentially means that a set of triples is generated from an image, and there is no checking of semantic consistency after semantic decoding.
\begin{myhyp}
In our model, and early on in brain evolution and development,
semantic decoding of sensory input behaves like a chatterbox.
\end{myhyp}
Chatterbox decoding has many advantages. E.g., it is fast and it informs the agent about the uncertainty in the decoding by providing several optional interpretations.


\begin{myhyp}
An open question is what kind of reasoning is performed in the mind already at the representation level (system-1)
and how much is performed at the explicit sentence level (system-2)?
 The latter might require executive functions, in particular in the case of ontological reasoning.
\end{myhyp}

\subsection{Semantic Memory}
\label{sec:semantic}

The probabilistic  KG can be related to the brain's semantic memory.
In visual perception, semantic memory defines the prior distribution for observing triples.
A second function is that semantic memory informs the agent about non-perceptual background.


A biological function is that semantic memory learns from the past to contribute to the present.
It fills in background knowledge about the non-perceived world by 
generating triples from the perceptual semantic memory and
from the  non-perceptual background memory:
Even if in the image the lion looks sleepy and friendly, lions are often aggressive (perceptional semantic memory) 
and  lions are dangerous beasts (background  semantic memory).

In  literature, semantic memory is sometimes equated with concept engrams \cite{ralph2017neural}. From our viewpoint, there is a clear distinction. Concept engrams are the basis for semantic decoding, employed in perception, episodic and semantic memory. What is special about the semantic memory in our model
 is the semantic memory embedding vector which is the engram specific to semantic memory.
\begin{myhyp}
Both in our model and the brain, perceptual semantic memory emerged out of perception's semantic decoder.
In addition to the concept engrams, our semantic memory model requires the time-invariant semantic memory embedding $\mathbf{\bar a}$ (Equation~\ref{eq:subs}), which could be related to the perceptual semantic memory engram in the brain and which might be attached with the semantic index $\bar e$.
See Figures~\ref{fig-arch} and~\ref{fig-Architecture}.
\end{myhyp}


In our model, both perception and semantic memory rely essentially only on the semantic decoder.
This might explain the robustness of semantic memory to brain damage and aging.
Only in the case that working memory deteriorates, both perceptual decoding and semantic memory should deteriorate, as well.
\begin{myhyp}
In both our model and the brain, semantic memory contributes to enrich perception by generating triple statements, a process which might be triggered through perception.
\end{myhyp}
If the agent sees a lion, it might want to recall that lions can be aggressive, without a recall of a specific past experience.

Semantic memory shows the power of semantic decoding. In our model, semantic memory is realized by a fixed vector $\mathbf{\bar a}$, so it appears to contain little information. In connection with the semantic decoder, the vector recovers all of semantic memory.

\subsection{Episodic Engrams and Memory}
\label{sec:episo}

In contrast to semantic memory, episodic memory requires a recollection of a prior experience \cite{tulving1985elements}.
The realization of an episodic memory engram
 in our model is straightforward: for each memorable event, e.g. attached with
 emotion or novelty, we introduce a time index, $e_t$.
 Then, its embedding is the activation of the representation layer at time instance $t$, i.e., $\mathbf{a}_t$,
 which becomes the
 connection weight vector between the time index and the representation layer (Equation~\ref{eq:visual}).
\begin{myhyp}
In our model, a new episodic engram for time instance $t$ consists of a novel time index $e_t$ and its embedding, i.e., its connection vector to the representation layer, which reflects $\mathbf{a}_t$. This is a greatly simplified mathematical model of the much more complex operations in the brain.
\end{myhyp}
For memory recall, the time index is activated, resulting in the activation of the representation layer with
the stored embedding vector $\mathbf{a}_t$.\footnote{We ignore here that $\mathbf{a}_t$ contains information on the three bounding boxes which need to be presented sequentially.}
\begin{myhyp}
An episodic memory recall consists of the reconstruction of the embedding of the episodic engram by activating the index $e_t$ for the past time instance $t$, which leads to an engram reconstruction in the representation layer. A similar reconstruction might happen already in animals' brains; subsequent declarative decoding, requiring a complex semantic decoder, might only be realized in humans.
\end{myhyp}
Thus the recall of a past episodic memory associated with $e_t$ happens primarily by a reconstruction of $\mathbf{a}_t$. Due to connectional reciprocity, this might activate sensory memory layer $\mathbf{g}$ and maybe even earlier processing layers restoring the sensory impression of the past episodic memory.
 Episodic memory storage and recall are on the representation level, which might already be relevant for animals, in general. Declarative decoding of a past episodic memory, in contrast, might be unique to humans: We can verbally report about past episodic memories!

In the mind, episodic memory is the result of rapid associative learning in that
a single episode and its context become associated and bound together and can be retrieved from memory after a single episode.
Episodic memory stores information
of general and personal events \cite{tulving1972episodic,tulving1985elements,tulving2002episodic,gazzaniga2004cognitive}
and concerns information we ``remember'' including the
spatiotemporal context of events \cite{gluck2013learning}.
In the mind, an episodic memory experience is an active process that involves
details of the event and its location \cite{moscovitch2016episodic}. Sometimes the reconstruction is considered a
Bayesian process of reconstructing the past as accurately as possible based on available engram information.

The formation  of indices for episodic memory in the hippocampus is a well-accepted theory \cite{tonegawa2018role}.
It goes back to the hippocampal memory indexing theory \cite{teyler1986hippocampal,teyler2007hippocampal}, which was long controversial.
The indices have a {relational memory function} in the sense that they bind together
different dimensions in the representation layer.
Little is known about how exactly new episodic indices are formed in the brain and how they quickly set up the connection patterns to the representation layer. Evidence for time cells in the hippocampus (CA1) have
recently been found \cite{eichenbaum2012towards,eichenbaum2014time,kitamura2015entorhinal,kitamura2015entorhinal2}.
There is some evidence that indices might be quickly formed in the hippocampus by a process termed neurogenesis.
Neurogenesis has been established in
the dentate gyrus (part of the hippocampal formation) from stem cells throughout adult life;
these new neurons may be preferentially recruited in the formation of memories.
In fact, it has been observed that the adult macaque monkey
forms a few thousand new neurons daily \cite{gluck2013learning,gould1999neurogenesis}, possibly
to encode new information \cite{becker2005computational}.

Retrieval of past episodic memory engrams can serve many purposes, discussed now.

\subsubsection*{1: Short and Medium Term Memory for Context}

Episodic memory can provide a short to medium term memory,
providing information which is not readily available by sensory input or other forms of short term memory:
As an example,
an agent should be aware why it is in the hide-out and that there is still a leopard waiting outside the hide-out,
even when the vicious beast is not visible.
This form of episodic memory is sometimes affected by aging, with deteriorations in both MTL and working memory \cite{daselaar2013age}. In general, the  mechanisms for forming new episodic memories are complex and thus quite sensitive to traumatic brain injury.

\subsubsection*{2: Replay for Training}

Episodic memory can be used to train implicit memories and semantic memory embedding, as well as decision modules during a consolidation step, possibly by replay during sleep. This is the basic idea behind the complementary learning systems (CLS) theory \cite{mcclelland1995there,kumaran2016learning}.

 A gradual transition from episodic to semantic memory can take place, in which episodic memory reduces its sensitivity and association to particular events, so that the information can be generalized as semantic memory.
 Without a doubt, semantic and episodic memories support one another \cite{greenberg2010interdependence}.
 Thus some theories speculate that episodic memory may be the
``gateway'' to semantic memory \cite{baddeley1974working,squire1987memory,baddeley1988cognitive,steyvers2004word,socher2009bayesian,mcclelland1995there,yee2014cognitive,kumar2015ask}.

In our model, this form of gradual learning concerns different components, first, the deep neural network which maps visual input to latent representation,
second, concept engrams and, third, the semantic memory engram.
Concept representations and the deep learning module might require little, if any, fine tuning.
Concept representations might have attractor properties in high dimensions, such that concept fine-tuning might be robust. Training the semantic memory engram $\mathbf{\bar a}$, in contrast, is important for modifying and extending semantic memory. Challenging is the absorption of new information without losing existing information (catastrophic forgetting). Semantic memory replay might be a solution, where triple samples generated by semantic
memory itself are used in self-supervised training, together with samples generated in perception.

\subsubsection*{3: Imagination and Planning}

Episodic memory permits future-oriented
mental time travel to evaluate future consequences of actions \cite{schacter2012future}. Humans are able to mentally place themselves in the past, in the future, or in counterfactual situations, a process called
autonoetic consciousness.

\subsubsection*{4: Consolidation in Long Term Memory for Decision Support}

Systems consolidation of memory (SCM) concerns the consolidation of memory into neocortex.
The \textit{standard theory} assumes that, at some point, episodic memory becomes independent of hippocampus and MTL over a period of weeks to years \cite{squire1995retrograde,frankland2005organization}.
In contrast, the\textit{ multiple trace theory} assumes that hippocampus and MTL remain involved \cite{nadel1997memory,jonides2008mind,greenberg2010interdependence}.
In general, it is assumed that consolidation involves both MTL and the medial prefrontal cortex (mPFC) \cite{tonegawa2018role}.

During the course of consolidation,
 memories need to become
interleaved into a network of existing related memories in
the neocortex \cite{edgell1929child,bartlett1995remembering}.
This interleaving process incorporates new
memories and typically requires modifications of the preexisting
network structure to add the new memories \cite{preston2013interplay}.

\begin{myhyp}
Following our mathematical model, systems consolidation of memory in the brain would involve a transfer
of time indices from MTL to the neocortex, possibly by replay,
where these index duplicates would inherit the connection weights. For a while, both representations exist in parallel, but gradually, the index representation in neocortex becomes dominant.
\end{myhyp}

Episodic memory will mostly consolidate in long term memory those events
that are memorable, unexpected or attached with emotion. Associated with those memories might be past decisions and actions, potentially with attached outcomes.
An important capability for an agent is the comparison of the current situation to previous experiences:
If a current event is very similar to a past event,
and that past event triggered a certain action,
it makes sense that the current event should trigger the same action ---if it led to a good outcome--- including a prognosis of what to expect next.
As an example: the agent finds the current situation very similar to a previous one where the next thing was an attack by a leopard, so better watch out!
A recall can simply be triggered based on similarities in the representation layer, potentially involving an attention mechanism. In general, much of the temporal resolution is lost in consolidation,
 and episodic memories become more like snapshots rather than detailed sequences of events. Memory consolation might be a process executed completely or partially during sleep \cite{stickgold2005sleep}.

\subsection{The Relationship between Semantic Memory and Consolidated Episodic Memory}

Some researchers consider semantic memory simply as being consolidated episodic memory. Our model would support a very close relationship between both. After all, the only difference in our model is that an episodic memory recall requires the activation of the consolidated time index $e_t$, whereas a semantic memory requires the activation of the semantic memory index $\bar e$.
The semantic engram, consisting of $\bar e$ and $\mathbf{\bar a}$, might be established directly in neocortex, and not first in hippocampus.
Mathematically, a recall of a semantic memory becomes close to the activation of all time indices concurrently, since
$\mathbf{\bar a} \approx 1/N_T \sum_{t=1}^{N_T} \mathbf{a}_t$, and, in this sense, a semantic memory is an episodic memory, where the time index is lost.
Thus one thing that is going on in memory consolidation during sleep is that we adjust our prior distribution to new experiences!



\section{Experiments}
\label{sec:exp}

\subsection{Dataset}
We tested our model on the \emph{Visual Relation Detection (VRD)} \cite{lu2016visual} dataset, which is the basis for many research works on visual relationship detection. We used the more commonly used split of this dataset which contains 100 object classes and 70 predicates with 4000 images used for training, and 1000 images for testing. These images contain an overall number of 37,993 triples from which 6,672 are unique, and 1,877 of these unique triples are in the test set for zero-shot evaluations.

\subsection{Metric}
We report our results using Recall@K metric which is defined as the mean ratio of ground truth labels in each image that appears in the model's top $K$ predictions. Recall@K is a popular choice in most of the related visual relation detection studies. The main reason is the incompleteness of visual relation detection datasets, i.e., some relations might not be annotated in the test set, while due to the model's generalization, they might get higher prediction values than the annotated ones.

\subsection{Architectures}
We use Faster R-CNN for the object detection backbone employed before the sensory memory layer. The VGG-16 architecture within this framework \cite{simonyan2014very} is pre-trained on ImageNet \cite{russakovsky2015imagenet} and fine-tuned to our data for relation detection. The sensory memory $\mathbf{g}$ is a fully connected layer with 4096 neurons. Representation layer $\mathbf{q}$ and working memory layer $\mathbf{h}$ are also fully connected layers with 4096 and 500 neurons respectively.

\subsection{Results}
Table~1 shows the results of our ablation studies and also from \cite{baier2017improving}.
Unless specified otherwise, we report Recall@100. Here, ph stands for phrase detection and pr stands for predicate detection. In phrase detection setting we are concerned with the detection rate of both the bounding boxes and the triples. In predicate detection, subject concept and object concept are both given and the task is to predict the predicate.
For z-s-ph/z-s-ph (zero-shot), we only evaluate the test set performance on triples that did not occur
in training. The first row (Model) shows results for our model. Our model gives better results for the zero-shot experiments. The last two columns report recall results for only the semantic memory (Recall@10, Recall@1). The first row shows results where the semantic memory was extracted from our perceptual model. The results (82.46 and 53.53) are worse than the result for \cite{baier2017improving}, where the latter was trained directly on the triple data. S1 and S9 show results where we added 1 and 9 epochs of pure semantic training to the perception model. We see that the semantic model improves significantly with almost no degradation on perception. Dir are results where we removed the working memory, i.e., we used the simple decoder.


\begin{table}[t]
 \begin{center}
 \caption{Experimental results.
}
 \begin{tabular}{|l|c|c|c|c|c|c|}
\hline
Method & ph & z-s-ph & pr & z-s-pr & @10 & @1 \\
\hline \hline
Model & 23.45 & \textbf{10.95} & 93.32 & 78.79 & 82.46 & 53.53 \\
S1 & 23.32 & 10.44 & 93.17 & \textbf{80.07} & 93.46 & 67.55 \\
S9 & 22.61 & 9.24 & 92.77 & 79.47 & 94.77 & 68.68 \\
Baier & \textbf{25.11} & 7.96& \textbf{93.81} & 76.05 & \textbf{95.86} & \textbf{70.50} \\
Dir & 11.13 & 7.87 & 77.19 & 65.61 & - & - \\
Rand & 0.01 & 0.00 & 18.53 & 16.51 & 0.08 & 0.01 \\
\hline
\end{tabular}
 \end{center}
 \label{tab:res}
\end{table}

\section{Conclusions and Future Work}
\label{sec:concl}

We have presented a mathematical model for perception, episodic memory,  and semantic memory, and related it to cognitive models of the human brain.
Our mathematical approach is based on tensors and their embedding models. In particular, we suggested that indices and index embeddings, which are central to tensor models, are equally important for an understanding of the functioning of the human brain.
We have shown how knowledge graphs can be queried by a stochastic triple sampling process, which might be related to the way that the brain queries memories.


Our main hypothesis is that episodic memory, semantic memory, and to some degree also working memory, are
by-products of the need for humans to extract more meaningful and more complex information from sensory inputs.
For the generation of explicit triple statements, memory and perception share a semantic decoder.
We have discussed the close relationship between semantic memory and episodic memory.


Attention is an important topic to explore in future work. Attention already plays a big role in our approach
but its real role is much greater. First, the selection of the bounding boxes for subject, object, and predicate contains an attention process. By focussing on selected bounding boxes, we greatly reduced complexity and avoided the challenging task of finding a scene explanation by multiple causes, a difficult issue
 that needs to be addressed in the Harmonium \cite{smolensky1986information} and the restricted Boltzmann machine \cite{hinton2010practical}.
Second, an attention mechanism might guide the sampling into the \textit{s-o-p} order, i.e., in sampling an entity first (subject), then another entity (object), and finally a predicate.
 Third, we can generate triples either from perception, or from episodic memory, or from semantic memory.
 The decision,  in which situation  which one of these three sources should generate samples to excite the representation layer, also requires an attention mechanism.


A second future project considers forecasting. Forecasting in temporal knowledge graphs is an active area of research and certainly is relevant for understanding the mind's operations, in particular in the context of a Bayesian brain theory \cite{schultz1997neural}.

It would also be interesting to explore links to decision making and action initiation. Improving
both is the ultimate goal and is essential for an agent's survival \cite{wolpert2001perspectives}.
Decision making is associated with the executive functions of the working memory. In our model, inputs for the decision modules would be provided by working memory
and by the representation layer.
 In principle, units in the index layer could also be relevant,
 but since an index activates many units in the representation layer, the latter is likely to dominate the response.
In our model, decisions can be made very quickly as a direct reaction to visual input.
On the other hand, a decision based on the semantic decoding of the complete scene would be quite involved and would take longer, but might be more informed. Episodic information can recover past experiences and their emotional attachments and thus influence decisions, as can input from semantic memory.
Another interesting view is that decisions are made without involving conscious decision making: \cite{gazzaniga2004cognitive} suggests that consciousness is a tale that the brain tells itself to explain what it is doing, by using the so-called left-brain interpreter.

Finally, it would be interesting to explore the question of how the brain manages to quickly store a new episode. As discusses, a technical implementation is quite simple: A new index is formed, together with a connection vector copying the episodic memory trace. How this is done in brainware is still largely unknown~\cite{quiroga2013brain}.

Model performance can certainly be improved in many ways, but our aim was the simplest plausible model obeying basic constraints imposed by brainware.

This paper was concerned with the question:
Why did humans develop semantic decoding and explicit memory in the first place, and what are relationships to technical systems? The focus was not on the study of the fully developed faculties of the human mind, with its
amazing dimensions of intellectual skills.


\subsubsection*{Note Added in Proof}

The experimental results are preliminary and will be updated shortly. Code will be made available, as well.

\bibliographystyle{apacite}

\setlength{\bibleftmargin}{.125in}
\setlength{\bibindent}{-\bibleftmargin}

\bibliography{TensorBrainSem}

\section{Appendix: Semantic Decoding}

We describe a concrete implementation of the triple generating process for perception described in Section~\ref{sec:serial}, under the architectural constraints discussed in Section~\ref{sec:bioImp}.

The three bounding boxes are sequentially presented as visual inputs.
A bounding box triple defines a time instance $t$. Figure~\ref{fig-Architecture} (top) shows the processing steps.
Figure~\ref{fig-Architecture} (bottom) shows the processing steps as suggested in ~\cite{trespmodel2019}}.

\subsection{Predicting the Subject Identity}

Consider first $\textit{BB}_{\textit{sub}, t}$. In the first step, the activation in the representation layer is
\[
\mathbf{q} := \mathbf{a}(\textit{BB}_{\textit{sub}, t})
\]
where
$
\mathbf{a}(\textit{BB}_{\textit{sub}, t}) = D \times \mathbf{g}(\textit{BB}_{\textit{sub}, t}) .
$
Here, $\mathbf{g}(\textit{BB}_{\textit{sub}, t})$ describes a deep neural network mapping from
$\textit{BB}_{\textit{sub}, t}$ to $\mathbf{g}$,
 and $D$ is a learned connection matrix.
Then,
\[
\mathbb{P}(S=s | \textit{BB}_{\textit{sub}, t}, \textit{BB}_{\textit{pred}, t}, \textit{BB}_{\textit{obj}, t})
= \textrm{softmax}_S (\mathbf{ a}^T_s \mathbf{q}) .
\]
This represents the activation of the entity indices as a response to the visual input and corresponds to Equation~\ref{eq:ssi} in our tensor model.

In the next instance, we sample from this categorical distribution.
The sampled $s^*$ then activates the representation layer, as
$
\mathbf{q} := \mathbf{a}_{s^*}
$,
and the whole mind is informed about the sampled subject entity.\footnote{Note that, for biological reasons (we cannot look into the future), we have made the simplification that
\[
\mathbb{P}(S=s | \textit{BB}_{\textit{sub}, t}, \textit{BB}_{\textit{pred}, t}, \textit{BB}_{\textit{obj}, t})
\approx
\mathbb{P}(S=s | \textit{BB}_{\textit{sub}, t})
\]
This can be justified by assuming that the subject is well defined by the subject bounding box.}
Technically, the mapping from $\textit{BB}_{\textit{sub}, t}$ to the indices is a feedforward neural network classifier
with $N_E$ outputs.\footnote{Written in the form of Equation~\ref{eq:ssi},
$\textit{f}^{\textit{sub}}_{\mathbf{w}}(\mathbf{a}_s, \mathbf{a}_t) = \mathbf{a}_s^T \mathbf{a}(\textit{BB}_{\textit{sub}, t})$.}

\subsection{Predicting the Object Identity}

Information about
the subject sample $s^*$ and its visual grounding in the current bounding box, $\mathbf{a}(\textit{BB}_{\textit{sub}, t})$,
 is stored in working memory layer $\mathbf{h}$, as
\[
\mathbf{h}^S = \textrm{sig}
\left(
V
(
\mathbf{a}_{s^*} + \mathbf{a}(\textit{BB}_{\textit{sub}, t})
)
\right) .
\]
Here, $V$ is the connection matrix from the representation layer to the working memory layer.
We set
\[
\mathbf{q} := \mathbf{a}(\textit{BB}_{\textit{obj}, t}) + W \textrm{sig} (B \mathbf{h}^S)
\]
with
$
\mathbf{a}(\textit{BB}_{\textit{obj}, t}) =
D \times \mathbf{g}(\textit{BB}_{\textit{obj}, t})
$; $\mathbf{q}$ now represents sensory information from the object bounding box and input from the working memory.
$W$ is the connection matrix from the working memory to the representation layer.

An object $o^*$ is then sampled from
\[
\mathbb{P}(O=o|S=s^*, \textit{BB}_{\textit{sub}, t}, \textit{BB}_{\textit{pred}, t}, \textit{BB}_{\textit{obj}, t})
= \textrm{softmax}_O (\mathbf{ a}^T_o \mathbf{q}) .
\]
This represents the activation of the entity indices as a response to the visual input and subject information and corresponds to Equation~\ref{eq:soi} in our tensor model.
The sampled $o^*$ then activates the representation layer, as
$
\mathbf{q} := \mathbf{a}_{o^*}
$,
and the whole mind is informed about the sampled object entity.\footnote{Here we approximate,
\[
\mathbb{P}(O=o | S=s^*, \textit{BB}_{\textit{sub}, t}, \textit{BB}_{\textit{pred}, t}, \textit{BB}_{\textit{obj}, t})
\approx
\mathbb{P}(O=o | S=s^*, \textit{BB}_{\textit{sub}, t}, \textit{BB}_{\textit{obj}, t}) .
\]
Written in the form of Equation~\ref{eq:soi},
\[
\textit{f}^{\textit{obj}}_{\mathbf{w}}(\mathbf{a}_{s}, \mathbf{a}_o, \mathbf{a}_t) = \mathbf{a}_o^T \left(\mathbf{a}(\textit{BB}_{\textit{obj}, t})
+ W \textrm{sig} (B \textrm{sig} (V (\mathbf{a}_{s} + \mathbf{a}(\textit{BB}_{\textit{sub}, t})))) \right) .
\]
}

\subsection{Predicting Predicates}

In the third step, we predict the predicates.
We aggregate all past information into the working memory layer, as
\[
\mathbf{h}^{S, O} =
\textrm{sig}
\left(
V
(\mathbf{a}_{o^*} + W \textrm{sig} (B \mathbf{h}^S) + \mathbf{a}(\textit{BB}_{\textit{obj}, t}))
 \overbrace{+ B \mathbf{h}^{S}}^{optional}
\right) .
\]
We set
\[
\mathbf{q} := \mathbf{a}(\textit{BB}_{\textit{pred}, t}) + W \textrm{sig}(B \mathbf{h}^{S, O})
\]
with
$
\mathbf{a}(\textit{BB}_{\textit{pred}, t}) =
D \times \mathbf{g}(\textit{BB}_{\textit{pred}, t})
$; $\mathbf{q}$ now represents sensory information from the predicate bounding box and input from the working memory.
We get
\[
\mathbb{P}(P=p | S=s^{*}, O={o^*}, \textit{BB}_{\textit{sub}, t}, \textit{BB}_{\textit{obj}, t}, \textit{BB}_{\textit{pred}, t})
= \textrm{softmax}_P (\mathbf{a}^T_p \mathbf{q}) .
\]
This represents the activation of the predicate indices as a response to the visual input and corresponds to Equation~\ref{eq:spi} in our tensor model.


The sampled $p^*$ then activates the representation layer, as
$
\mathbf{q} := \mathbf{a}_{p^*}
$,
and the whole mind is informed about the sampled predicate entity.\footnote{It is important that subject and object embeddings enter asymmetrically in the equation,
so the agent can distinguish between ``dog bites person'' and ``person bites dog''. Written in the form of Equation~\ref{eq:spi},
\[
f^{pred}_{\mathbf{w}}(\mathbf{a}_s, \mathbf{a}_p, \mathbf{a}_o, \mathbf{a}_t)
= \mathbf{a}_p^T
\left(
\mathbf{a}(\textit{BB}_{\textit{pred}, t})
+ W \textrm{sig}
(
B \textrm{sig}
[
V
(\mathbf{a}_{o} + W \textrm{sig} (B \textrm{sig}
(V
(
\mathbf{a}_{s} + \mathbf{a}(\textit{BB}_{\textit{sub}, t})
)) ) + \mathbf{a}(\textit{BB}_{\textit{obj}, t}))
\overbrace{+B \textrm{sig}
(V
(
\mathbf{a}_{s} + \mathbf{a}(\textit{BB}_{\textit{sub}, t})
)
)}^{optional}
]
)
\right) .
\]
}


\subsection{Integrated Triple Representation}
\label{sec:inttrip}

After decoding, we can calculate
\[
\mathbf{h}^{S, P, O} = \textrm{sig}
\left(
V(\mathbf{a}_{p^*} + W \textrm{sig} (B \mathbf{h}^{S, O})
+ \mathbf{a}(\textit{BB}_{\textit{pred}, t}))
\overbrace{+B \mathbf{h}^{S, O}}^{optional}
\right)
\]
and
\[
\mathbf{q} := W \textrm{sig}(B \mathbf{h}^{S, P, O})
\]
which are triple statement embeddings. They integrate information about the triple and can also serve as an initialization for follow-up triple decoding, providing context.

\subsection{Interpreting Probabilities and Samples}

We review the  discussion presented in  the main part of the  paper. 


\subsubsection*{Perception and Episodic Memory}

Assuming a sufficiently well performing decoding system,
any decoded triple sample $(s^*, p^*, o^*)$ represents very likely
a true triple statement, in the sense that $y_{s, p, o, t} =1$. Example: at time instance $t$, we decode that
\textit{(Jack, looksAt, Mary)}.   Following the above formulas, we should get
\[
\mathbb{P}(S=s, P=p, O=o | \textit{BB}_{\textit{sub}, t}, \textit{BB}_{\textit{pred}, t}, \textit{BB}_{\textit{obj}, t}) \approx \frac{1}{\sum_{s, p, o} y_{s, p, o, t}}
\]
which, even for true triples,  can be much smaller than $1$. Thus the samples are easier to interpret than the conditional probabilities.

In sampling,  we can fix  a particular $\tilde s$ as being the subject, instead of using a sampled $s^*$. Then the sampling would produce a
 decoded triple sample $(\tilde s, p^*, o^*)$; i.e.,  conditioning results in a selection process concerning the true triple statements. As an example, we might have selected $\tilde s = \textit{Mary}$ and we decoded   that
 \textit{(Mary, looksAt, David)}.


\subsubsection*{Semantic Memory}

Any decoded triple sample $(s^*, p^*, o^*)$ represents  a sample from the prior distribution for the perception process. Thus it should be interpreted as the prior probability that a triple is generated in the perceptual sampling process. The semantic sampling might have produced that $\textit{(Jack, looksAt, Mary)}$, which is a probable triple in the prior distribution.   As before, we might have selected $\tilde s = \textit{Mary}$ and we decoded   that often
 \textit{(Mary, looksAt, David)}.

As discussed in Section~\ref{sec:trainsm}, we assume that the background probabilistic  KG is a static KG. With a perfect model, sampling from this  static KG produces with high likelihood only true triples. As example: a triple produced by the background semantic memory might be that $\textit{(Jack, sibling, Mary)}$.
As before, we might have selected $\tilde s = \textit{Mary}$ and we decoded   that 
 \textit{(Mary, sibling, Jack)}. With incomplete
  data and an approximating model, of course, true triples might be missed in sampling.

\begin{figure}[t]
\vspace{-1cm}
\begin{center}
\includegraphics[width=1.0\linewidth]{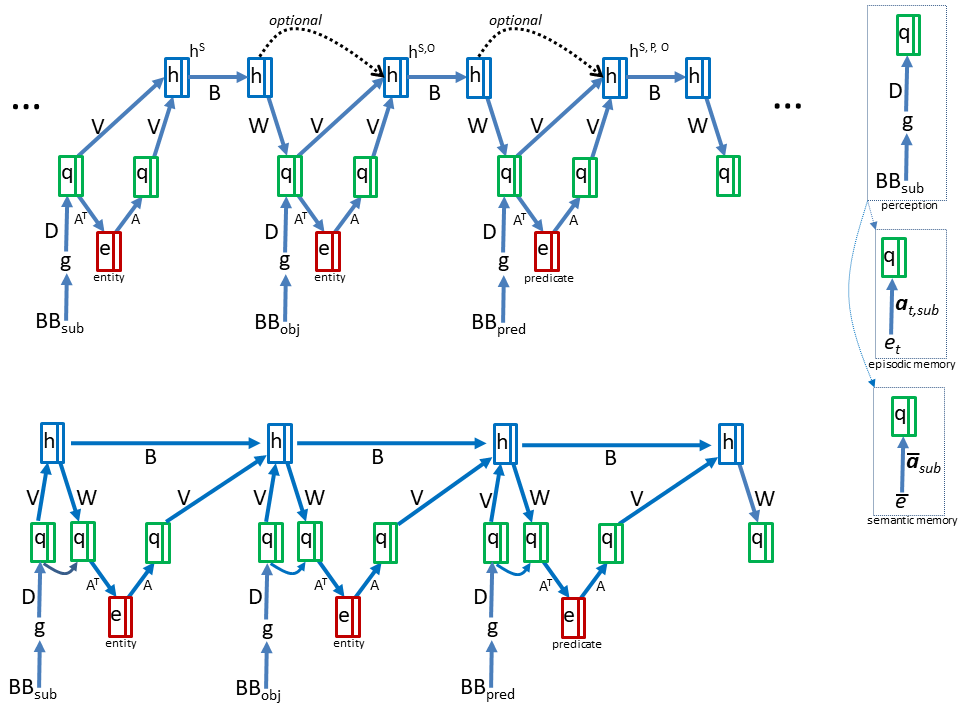}
\end{center}
\vspace{-0.5cm}
\caption{
Perceptual decoding. Each node represents a complete layer.
The $\mathbf{q}$-layer units (green) have a linear response,
the $\mathbf{h}$-layer units (blue) have a sigmoidal response, and the $\mathbf{e}$ layer (red) has a joint softmax response, from which an index is sampled.
Top: Operations as described here.
The architectures implement several constraints:
First, all communication between layers passes through the representation layer $\mathbf{q}$.
Second, there is a direct path $\textit{BB} \rightarrow \mathbf{g}\rightarrow \mathbf{q}\rightarrow \mathbf{e}$ for fast perception.
Third, the working memory layer $\mathbf{h}$ layer integrates all previous information (information on present and past visual inputs and on past sampling decisions). Fourth, via $\mathbf{q}$, the index layer $\mathbf{e}$ receives information from the bounding box $\textit{BB}$ and the working memory layer $\mathbf{h}$.
The three boxes on the right side indicate the replacement of the sensory module by episodic memory, resp., semantic memory; shown here for the subject, but similar for object and predicate. Note, that for episodic memory we have the decomposition
$
 \mathbf{a}_t = [\mathbf{a}(\textit{BB}_{\textit{sub}, t}); \mathbf{a}(\textit{BB}_{\textit{pred}, t}); \mathbf{a}(\textit{BB}_{\textit{obj}, t}) ]
 $
and
for semantic memory
$\mathbf{\bar a} = [\mathbf{\bar a}_{\textit{sub}}; \mathbf{\bar a}_{\textit{pred}}; \mathbf{\bar a}_{\textit{obj}}]$.
The dots to the left and right side indicate that complex decoding can be chained and information can be propagated forward to the next triple decoding.
Bottom: Alternative implementation.
}
\label{fig-Architecture}
\end{figure}

\end{document}